\documentclass[journal]{IEEEtran}
\usepackage{amsmath}
\usepackage{mathtools}           % 提供 \coloneqq 等，必须在 amsmath 之后
\usepackage{newtxtext,newtxmath} % 替换默认字体为 Times 风格 (会自动处理 amssymb/amsfonts，避免冲突)

\usepackage{graphicx}
\usepackage[caption=false,font=footnotesize]{subfig} 
\usepackage{stfloats}            % 允许双栏环境下的跨栏图片出现在底部
\usepackage{placeins}            % 提供 \FloatBarrier 防止图片乱跑

\usepackage{array}
\usepackage{booktabs}            % 提供 \toprule, \midrule 等三线表命令
\usepackage{multirow}
\usepackage{makecell}
\usepackage{siunitx}             % 科学计数和单位排版

\usepackage{algorithmic}
\usepackage{textcomp}
\usepackage{url}
\usepackage{verbatim}

\usepackage[hidelinks]{hyperref}
\newcommand{\mcell}[1]{\textbf{#1}}

\renewcommand{\arraystretch}{1.15} % 行距更舒服

\def\BibTeX{{\rm B\kern-.05em{\sc i\kern-.025em b}\kern-.08em
    T\kern-.1667em\lower.7ex\hbox{E}\kern-.125emX}}
\usepackage{balance}
\begin{document}
\title{Stabilizing Neural-to-Image Retrieval with Frequency-Domain Priors and Adaptive Supervision}
\author{YaoYang Liu, ZhuoShi Tang, Haopeng Lai,  Weijun Sun*, Xiaozhao Fang*, Jie Wen, Fellow, IEEE%
	\thanks{YaoYang Liu, ZhuoShi Tang, Haopeng Lai and Weijun Sun and Xiaozhao Fang are with the School of Automation, Guangdong University of Technology, Guangzhou 510006, China (email: yaoyangliu2026@163.com, tangchaoshi@mails.gdut.edu.cn, HPLai\_GDUT@163.com, gdutswj@gdut.edu.cn, xzhfang168@126.com).}
	\thanks{Jie Wen is with the School of Computer Science and Technology, Harbin Institute of Technology, Shenzhen 518055, China (e-mail: jiewen\_pr@126.com).}
	\thanks{Corresponding authors: Weijun Sun(gdutswj@gdut.edu.cn), ORCID: 0000-0002-2342-4434; Xiaozhao Fang(xzhfang168@126.com)}
}
\maketitle

\begin{abstract}
Visual neural decoding aims to infer perceived visual stimuli from brain activity, offering important insights into human visual cognition while enabling transformative applications in brain computer interfaces and artificial intelligence. Recent advances in pretrained vision-language models have provided a semantically rich representation space for this task. However, reliable alignment between EEG/MEG signals and visual embeddings remains challenging because neural responses are noisy, highly variable across trials, and subject dependent. To address this issue, we propose a unified neural-to-image retrieval framework that improves retrieval performance by aligning neural and image representations in a shared semantic space. The framework integrates dual-stream adaptation, frequency-domain calibration, and phase-guided temperature modulation to promote more robust cross-modal semantic alignment. Specifically, dual-stream adaptation models a low-dimensional shared subspace with correlation regularization to suppress redundant noise and enhance cross-modal consistency. Frequency-domain calibration conditionally modulates neural representations using amplitude and phase statistics. Phase-guided temperature modulation further adjusts the supervision strength in contrastive learning according to the stability of neural responses. Experiments on THINGS-EEG and THINGS-MEG under both intra-subject and inter-subject settings demonstrate that the proposed framework consistently outperforms strong baselines. These results suggest that robust neural-to-image retrieval under heterogeneous neural conditions benefits from a stable shared semantic space, frequency-informed representation calibration, and adaptive supervision.
 Code will be released at \href{https://github.com/wanfkuabc/Geometry-Regularized-Brain-Embeddings}{github.com/wanfkuabc/Geometry-Regularized-Brain-Embeddings}.
\end{abstract}

\begin{IEEEkeywords}
Cross-Modal Retrieval, Brain-Computer Interface, Contrastive Learning, Low-Dimensional Adaptation, Visual Neural Decoding, CLIP.
\end{IEEEkeywords}

\section{Introduction}

\IEEEPARstart{D}{ecoding} visual semantic content from brain activity has become an increasingly important problem in neural decoding and multimodal learning~\cite{hebart2023thingsdata}. In neural-to-image retrieval, the goal is to identify images whose semantic content is consistent with the neural responses elicited during visual perception. Beyond its practical relevance to brain-computer interfaces, this task also offers a computational perspective on how the brain represents and organizes visual information~\cite{du2023decoding,ferrante2024decoding,li2024visual}. Among available recording modalities, EEG and MEG are particularly attractive because they provide millisecond-level temporal resolution, relatively low acquisition cost, and flexible experimental deployment. These advantages make them well suited for studying the fast dynamics of visual cognition and for building scalable neural decoding systems that go beyond small closed set recognition settings~\cite{deshpande2017new,orban2022review,baillet2017meg,gross2019meg}. At the same time, the fine temporal structure captured by EEG and MEG suggests that they may contain rich information about evolving perceptual representations, provided that this information can be translated into a semantic space suitable for retrieval.

Recent progress in large-scale vision-language models, especially CLIP, has created a promising foundation for retrieval-based neural decoding~\cite{du2023decoding,li2024visual,radford2021learning,nice60,things_meg,wu2025bridging}. Instead of formulating neural decoding as conventional closed set classification, CLIP-based neural-to-image retrieval maps neural responses into a pretrained visual embedding space and performs matching in a shared semantic space. This paradigm is attractive for several reasons. First, the pretrained visual space carries category-level and semantic structure learned from massive image-text supervision, which is difficult to obtain from neural data alone. Second, retrieval in such a space naturally supports zero-shot evaluation and cross-category generalization, which are much closer to the open-vocabulary setting of real-world perception~\cite{li2024visual,radford2021learning}. Third, using a frozen visual backbone allows the model to inherit a stable semantic reference while concentrating learning capacity on the neural side. However, the success of this paradigm ultimately depends not only on the quality of the pretrained visual space, but also on whether noisy neural observations can be mapped into that space in a sufficiently stable and transferable manner.

This alignment step remains the main bottleneck of current neural-to-image retrieval systems. Most existing methods follow a standard recipe that combines a pretrained visual encoder, a neural encoder, and a contrastive objective in a shared embedding space~\cite{li2024visual,nice60,singh2024learning,wu2025bridging}. Although effective, this formulation often treats alignment as a largely uniform projection problem and applies similar supervision across samples, batches, and subjects. For EEG and MEG, such a static treatment is often too coarse~\cite{li2024visual,nice60,wu2025bridging,singh2024learning,alghamdi2025cross,zhang2023subject}. Neural responses are not only noisy, but also highly variable across trials and exhibit substantial distribution shifts across subjects. As a result, the learned neural embeddings can fluctuate substantially in feature scale, dimension usage, and geometric organization across mini-batches, which in turn makes similarity estimation less reliable. Under these conditions, simply increasing encoder complexity does not directly solve the core problem, because the difficulty lies less in raw representation capacity than in the stability of the alignment pathway itself. What is needed is a mechanism that conditions neural representations before and during contrastive learning so that they can be organized more consistently with pretrained visual embeddings, while remaining robust to heterogeneous neural response quality.

Motivated by this view, we propose a unified framework for neural-based neural-to-image retrieval. Our method keeps the CLIP visual semantic space frozen and improves retrieval by conditioning the neural pathway at several complementary levels. Specifically, we introduce dual-stream adaptation and batch-wise stabilization to regularize the geometry of neural embeddings before cross-modal matching. We further incorporate frequency-domain priors by extracting amplitude and phase statistics from neural signals and using them to calibrate the learned representation through conditional modulation. In addition, we derive a phase-based control signal to adapt the contrastive temperature during training, allowing the supervision strength to respond to differences in neural response stability without requiring an explicit difficulty estimator~\cite{wang2020understanding,nunez2006electric,nunez1997eeg,liang2025adaptive,huang2025self,chaddad2023electroencephalography}. 

Our contributions are summarized as follows:
\begin{itemize}
	\item We propose a parameter-efficient alignment framework for neural-to-image retrieval. By combining lightweight dual-stream adaptation, batch-wise standardization, and low-dimensional shared-subspace regularization, the framework establishes a robust alignment pathway into a frozen CLIP visual semantic space.

	\item We introduce a frequency-domain enhancement branch that explicitly extracts amplitude and phase statistics from EEG and MEG signals and injects them into the shared semantic space through conditional modulation. In addition, we derive a phase-stability-based control signal to support adaptive supervision. This signal is used to modulate optimization under noisy neural conditions, rather than to explicitly estimate per-sample retrieval difficulty.
	
	\item We validate the proposed framework through extensive experiments on the THINGS-EEG and THINGS-MEG benchmarks. The results demonstrate consistent improvements across datasets. On THINGS-EEG under the 200-way zero-shot retrieval protocol, our method achieves \mcell{63.40\%}/\mcell{86.85\%} Top-1/Top-5 accuracy in the intra-subject setting, improving upon the previous best reported results by \mcell{12.5\%}/\mcell{7.15\%}. In the inter-subject setting, it attains \mcell{24.05\%}/\mcell{53.45\%} Top-1/Top-5 accuracy, outperforming a competitive baseline by \mcell{11.65\%}/\mcell{20.05\%}.
\end{itemize}

\section{Related Work}
This section reviews prior research most relevant to our problem from four perspectives: neural visual decoding and retrieval, shared brain-visual semantic spaces, parameter-efficient adaptation and alignment stabilization, and frequency-domain modeling with adaptive supervision. Rather than providing a purely taxonomic survey, we emphasize how these lines of work motivate the need for a retrieval framework that remains stable under noisy and heterogeneous EEG/MEG conditions.
\subsection{Visual Decoding and Retrieval from Neural Signals}

Decoding visual content from brain activity has long been a central topic in brain-computer interfaces and computational neuroscience~\cite{hebart2023thingsdata,ferrante2024decoding}. Early studies mainly focused on fMRI because of its high spatial resolution, but its low temporal resolution and limited experimental flexibility constrain its ability to characterize fast visual dynamics. By contrast, EEG and MEG provide millisecond-level temporal measurements and are therefore more suitable for studying the temporal evolution of visual perception~\cite{shi2025brainalign,hamalainen1993magnetoencephalography}. On this basis, a broad range of models, including convolutional architectures~\cite{craik2019deep,roy2019deep,schirrmeister2017deep}, temporal modeling networks, and Transformer-based designs, have been developed for EEG and MEG analysis, with encouraging results in image recognition, category classification, and visual reconstruction~\cite{li2024visual,nice60,schirrmeister2017deep,lawhern2018eegnet,fu2025brainvis,song2023decoding}.

The setting becomes more demanding when the task shifts from closed set prediction to open vocabulary or zero-shot image retrieval. In retrieval, neural signals must support fine-grained matching within a large candidate set rather than only coarse category discrimination. This change places greater emphasis on semantic structure, representation transferability, and robustness to noise. Compared with direct label prediction, retrieval-based formulations benefit from external semantic spaces that can organize candidate images at the representation level. Existing studies have therefore moved from conventional neural classification toward retrieval-oriented formulations that emphasize semantic correspondence across modalities. These developments indicate that modern neural decoding research increasingly depends on representation spaces that support both semantic generalization and robust matching.

\subsection{Brain-Visual Shared Semantic Spaces and Cross-Modal Alignment}

Motivated by the success of CLIP-based vision-language representation learning, recent work has explored shared brain-visual semantic spaces in which EEG or MEG responses are projected into pretrained visual embedding spaces~\cite{li2024visual,nice60,things_meg,wu2025bridging}. In this line of research, the visual encoder serves as a semantic reference learned from large-scale image-text data, while the neural encoder is trained to map brain responses into the same space. Relative to classification models trained only with neural supervision, this strategy provides richer semantic structure and naturally supports zero-shot retrieval as well as cross-category generalization.

Most existing methods optimize this shared space with contrastive objectives that pull matched neural image pairs together and separate mismatched pairs. Recent studies have also explored alternative brain-visual alignment strategies beyond standard pairwise matching, such as relational cross-modal alignment for visual classification~\cite{liu2025brain}. This design has substantially improved neural-to-image retrieval
for both EEG and MEG data and has established cross-modal
alignment as a dominant paradigm in the field~\cite{li2024visual,nice60,singh2024learning,wu2025bridging}. At the same time, many current formulations still rely on a relatively uniform projection rule and a shared similarity shaping strategy across heterogeneous neural samples. Since EEG and MEG data are noisy, nonstationary, and subject-dependent, the quality of alignment may vary considerably across trials and subjects, even when the visual space itself is strong. This suggests that preserving alignment quality under heterogeneous neural conditions remains a central challenge in shared brain-visual spaces~\cite{xu2025neurovista}.

\subsection{Parameter-Efficient Adaptation and Stable Alignment}

In brain-visual cross-modal learning, the visual branch is usually initialized from a large pretrained model, whereas the neural branch is learned from comparatively small and variable EEG or MEG datasets. Under this data regime, directly fine-tuning the full visual backbone can be computationally expensive and may also weaken the generic semantic structure acquired during large-scale pretraining. For this reason, parameter-efficient transfer strategies have become increasingly relevant~\cite{houlsby2019parameter}. Representative approaches include adapters, low-dimensional adaptation~\cite{hu2022lora}, and other lightweight trainable modules that preserve the frozen backbone while adding a limited number of task-specific parameters~\cite{sung2022vl}.

Alongside efficient transfer, recent studies have also paid increasing attention to the stability of shared embedding spaces. Standardization, normalization, low-dimensional compression, and correlation-based constraints have been used to reduce feature redundancy, alleviate distributional mismatch, and improve the organization of learned representations. These ideas are particularly relevant to EEG and MEG, whose statistical fluctuations can noticeably perturb the geometry of neural embeddings during training. Existing work has shown that lightweight adaptation and embedding regularization can each be beneficial. Related self-supervised representation learning methods, including BYOL~\cite{grill2020byol} and SimSiam~\cite{chen2020simsiam}, also highlight the importance of stabilization and redundancy reduction in learning transferable representations. But they are often introduced as separate components with distinct motivations. What remains unclear is how these stabilization and adaptation strategies should be organized within a unified retrieval framework.

\subsection{Frequency-Domain Modeling and Adaptive Supervision}

The frequency-domain structure of EEG and MEG signals contains physiologically meaningful information that is not always explicit in raw time-domain waveforms. Amplitude statistics describe the distribution of spectral energy and rhythmic strength across frequencies, whereas phase statistics reflect temporal synchronization, concentration, and response consistency, and are also closely related to functional connectivity analyses in EEG studies~\cite{anastasiadou2019graph}. For this reason, frequency-based representations have long been studied as informative descriptors for neural signal analysis~\cite{yi2022classification,rajagopal2025phase,stancin2021review}. In particular, circular phase statistics provide a compact way to characterize phase concentration and may offer useful cues about the stability of neural responses under noisy conditions.

A related line of research concerns adaptive supervision and optimization, where different samples are allowed to contribute unequally during training~\cite{zheng2025eeg}. Sample weighting, adaptive losses, and hardness-aware strategies have been explored in many learning settings, and these ideas are also relevant to cross-modal neural alignment because response quality can vary across trials and subjects~\cite{dominicus2025reliability,xu2024advancing}. In the current neural-to-image retrieval literature, frequency-based modeling and adaptive supervision have mostly been studied as separate themes. One emphasizes richer neural descriptors, whereas the other focuses on how training signals should be modulated across samples. To our knowledge, their interaction has been less systematically studied in retrieval-oriented neural alignment.

\begin{figure*}[t]
	\centering
	\includegraphics[width=\textwidth,trim=2mm 2mm 2mm 2mm,clip]{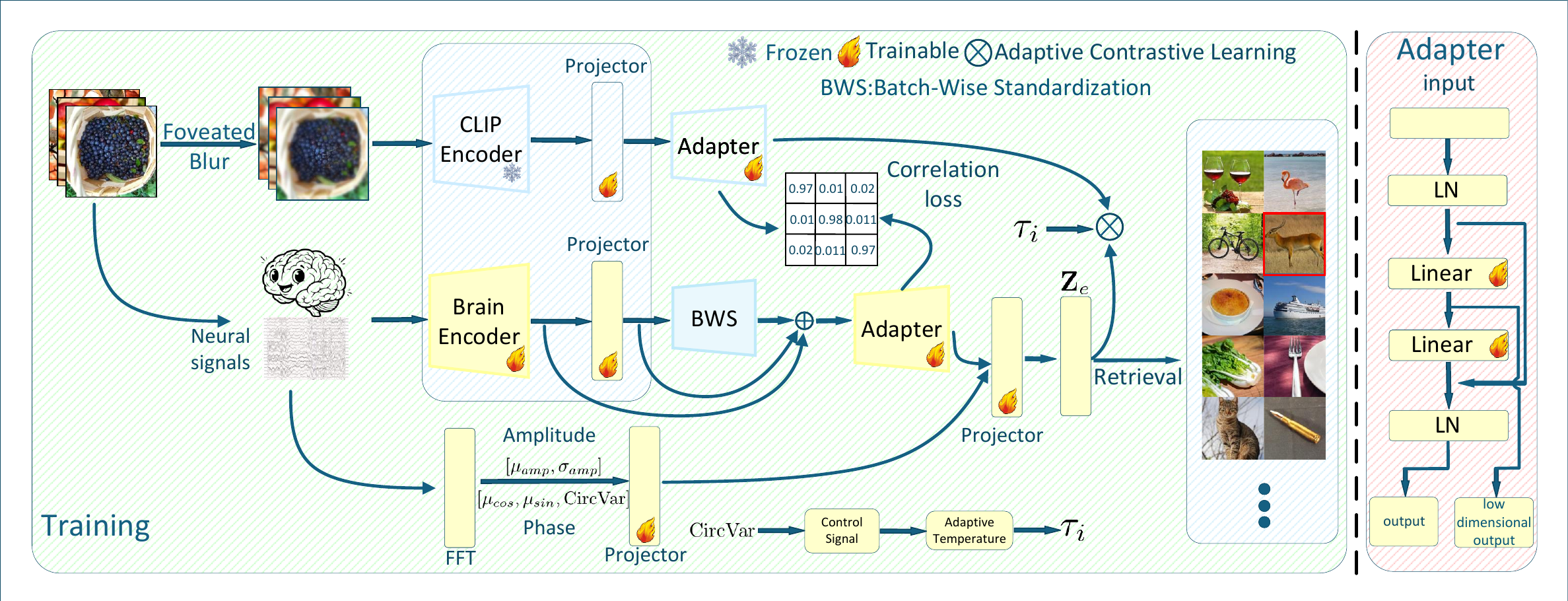}
	\caption{Overview of the proposed neural-to-image retrieval framework. \textbf{Left:} A frozen CLIP visual encoder first defines the target image semantic space, and the brain encoder maps neural signals into this shared space. To improve the stability of cross-modal alignment, the framework combines dual-stream adaptation, batch-wise standardization, and low-dimensional shared-subspace regularization. In parallel, a frequency-domain branch extracts amplitude and phase statistics from neural signals to conditionally calibrate neural embeddings. Phase-circular variance is further used to derive a phase-stability-based control signal, which generates sample-specific temperatures for adaptive contrastive learning. By jointly integrating frequency-based calibration, phase-based supervision control, and cross-modal structural regularization, the framework improves the robustness and retrieval performance of neural-to-image alignment. \textbf{Right:} Internal structure of the low-dimensional adapter. The adapter takes the input embedding, applies normalization and two lightweight linear transformations, and produces both the refined output and its low-dimensional bottleneck representation.}
	\label{fig:main-pipeline}
\end{figure*}

\section{Methodology}
\label{sec:method}
In this section, we describe the proposed neural-to-image retrieval framework from the overall formulation to its key alignment modules. We first define the retrieval problem and the neural encoding process, then introduce frequency-domain conditional modulation, batch-wise stabilization, and low-dimensional shared-subspace adaptation, and finally present the phase-guided adaptive contrastive objective used for training.
% =========================
% Notation & feature spaces
% =========================

\subsection{Problem Definition}
We study the neural-to-image retrieval problem, where the goal is
to identify the visual stimulus corresponding to a measured neural
response. Let $\mathcal{D}=\{(x_i,v_i)\}_{i=1}^N$ denote the paired
training set, where $x_i \in \mathbb{R}^{1 \times C \times T}$ is an
EEG/MEG sample recorded when a subject observes image $v_i$,
$C$ is the number of neural channels, and $T$ is the number of
temporal samples within the selected time window. For model
computation, the neural inputs are organized into a mini-batch tensor
$X \in \mathbb{R}^{B \times 1 \times C \times T}$, where $B$ denotes
the batch size. 
\subsection{Brain Encoding}
The neural encoder maps each EEG/MEG trial to a $d$-dimensional latent representation using a residual MLP projection network. Specifically, given a mini-batch $\mathbf{X}\in\mathbb{R}^{B\times1\times C\times T}$, we first reshape each sample into a vector and linearly project it into the latent space:
\begin{equation}
	\mathbf{h}_0 = \mathbf{W}_1 \mathrm{Flatten}(\mathbf{X}) + \mathbf{b}_1,
\end{equation}
where $\mathrm{Flatten}(\mathbf{X})\in\mathbb{R}^{B\times CT}$ denotes the vectorized neural input after removing the singleton dimension, and $\mathbf{W}_1$ and $\mathbf{b}_1$ parameterize the initial projection from the flattened neural input to the latent space.

To further refine the latent representation while preserving the lightweight design of the encoder, we apply a residual feed-forward transformation. The final neural embedding is defined as:
\begin{equation}
	\mathbf{z}
	=
	\mathrm{LayerNorm}\left(
	\mathbf{h}_0 + \mathbf{W}_3\big(\mathrm{Dropout}(\mathbf{W}_2 \mathbf{h}_0)\big)
	\right).
\end{equation}
where $\mathbf{W}_2$ and $\mathbf{W}_3$ parameterize the residual transformation for feature refinement, $\mathrm{Dropout}(\cdot)$ is used for regularization, and $\mathrm{LayerNorm}(\cdot)$ denotes layer normalization applied to stabilize the final embedding.

\subsection{Frequency-Domain Modulation with Amplitude and Phase Awareness}

To explicitly exploit the frequency-domain structure of neural signals, we introduce an amplitude- and phase-aware conditional modulation branch to recalibrate neural embeddings in the shared semantic space. Rather than directly concatenating frequency-domain features with the backbone representation, we treat amplitude and phase statistics as sample-specific priors and inject them into the neural embeddings through FiLM modulation~\cite{perez2018film}. This design allows the model to incorporate frequency-derived cues without altering the backbone architecture.
We first remove the singleton dimension and reshape the input as $\mathbf{X}\in\mathbb{R}^{B\times C\times T}$. We then apply a real-input fast Fourier transform along the temporal dimension to obtain a one-sided complex frequency representation:
\begin{equation}
	\mathbf{X}_{f}=\mathrm{rFFT}(\mathbf{X}),
\end{equation}
where $\mathbf{X}_{f}$ denotes the resulting complex representation in the frequency domain.

Based on $\mathbf{X}_{f}$, we compute the amplitude spectrum and phase spectrum as:
\begin{equation}
	\mathbf{A}=|\mathbf{X}_{f}|,\qquad
	\mathbf{\Phi}=\angle \mathbf{X}_{f},
\end{equation}
where $\mathbf{A}$ and $\mathbf{\Phi}$ denote the amplitude spectrum and phase spectrum, respectively.

To construct a compact frequency-domain descriptor, we extract five channel-wise statistics: the amplitude mean, the amplitude standard deviation, the phase cosine mean, the phase sine mean, and the phase-circular variance. For the 
$c$-th channel, these quantities are defined as follows:
\begin{equation}
	\mu_c^{a}=\frac{1}{F}\sum_{f=1}^{F}A_{c,f},\qquad
	\sigma_c^{a}=\sqrt{\frac{1}{F}\sum_{f=1}^{F}(A_{c,f}-\mu_c^{a})^2},
\end{equation}
where $A_{c,f}$ is the amplitude at the $f$-th frequency bin of the $c$-th channel, $\mu_c^{a}$ is the channel-wise amplitude mean, $\sigma_c^{a}$ is the channel-wise amplitude standard deviation, and $F$ is the number of frequency bins.

We next compute the first-order circular statistics of phase as:
\begin{equation}
	\mu_c^{\cos}=\frac{1}{F}\sum_{f=1}^{F}\cos(\Phi_{c,f}),\qquad
	\mu_c^{\sin}=\frac{1}{F}\sum_{f=1}^{F}\sin(\Phi_{c,f}),
\end{equation}
where $\Phi_{c,f}$ is the phase value at the $f$-th frequency bin of the $c$-th channel, and $\mu_c^{\cos}$ and $\mu_c^{\sin}$ denote the mean cosine and mean sine of the phase distribution, respectively.

Based on these quantities, the phase-circular variance is defined as:
\begin{equation}
	v_c^{\phi}
	=
	1-\sqrt{(\mu_c^{\cos})^2+(\mu_c^{\sin})^2+\epsilon},
\end{equation}
where $v_c^{\phi}$ denotes the phase-circular variance of the $c$-th channel, and $\epsilon$ is a small constant for numerical stability. These statistics jointly summarize spectral energy, dispersion, and phase concentration, providing compact sample-wise cues for frequency-conditioned modulation.

We then concatenate the channel-wise statistics to form a sample-level frequency descriptor:
\begin{equation}
	\mathbf{s}
	=
	\mathrm{Concat}\left(
	\boldsymbol{\mu_c}^{a},
	\boldsymbol{\sigma_c}^{a},
	\boldsymbol{\mu_c}^{\cos},
	\boldsymbol{\mu_c}^{\sin},
	\mathbf{v_c}^{\phi}
	\right),
\end{equation}
where $\boldsymbol{\mu_c}^{a}$, $\boldsymbol{\sigma_c}^{a}$, $\boldsymbol{\mu_c}^{\cos}$, $\boldsymbol{\mu_c}^{\sin}$, and $\mathbf{v_c}^{\phi}$ collect the corresponding channel-wise statistics across all channels. Under channel-wise aggregation, we obtain $\mathbf{s}\in\mathbb{R}^{5C}$. The descriptor is then mapped to a compact conditioning vector through a lightweight encoder:
\begin{equation}
	\mathbf{c}=f_{\mathrm{cond}}(\mathbf{s}),
\end{equation}
where $f_{\mathrm{cond}}(\cdot)$ consists of LayerNorm~\cite{ba2016layernorm}, a linear projection, and a GELU activation~\cite{hendrycks2016gelu}, and $\mathbf{c}$ denotes the resulting conditioning vector. This vector provides a compact summary of the frequency-domain characteristics of the current neural sample.

To inject these frequency-domain priors into the semantic representation, we employ a FiLM modulation module that predicts a scaling factor $\gamma\in\mathbb{R}^{d}$, a bias term $\beta\in\mathbb{R}^{d}$, and a confidence gate $q\in(0,1)$ from the conditioning vector $\mathbf{c}$. The scaling and bias parameters are generated by two independent MLP branches, whereas the confidence gate is produced by an additional branch followed by a sigmoid activation. To avoid over-modulating the backbone representation~\cite{he2016deep}, we constrain
the FiLM parameters to remain close to an identity perturbation:
\begin{equation}
	\gamma = 1 + 0.1\tanh(\hat{\gamma}),\qquad
	\beta = 0.1\tanh(\hat{\beta}),
\end{equation}
where $\hat{\gamma}$ and $\hat{\beta}$ are the raw outputs of the modulation network. 

Using the predicted modulation parameters, we first obtain the FiLM modulated representation:
\begin{equation}
	\tilde{\mathbf{z}}=\gamma\odot\mathbf{z}+\beta,
\end{equation}
where $\mathbf{z}$ denotes the original neural embedding, $\odot$ denotes element-wise multiplication, and $\tilde{\mathbf{z}}$ is the modulated neural embedding.

To further control the influence of frequency-derived conditioning, we introduce a confidence-controlled residual fusion mechanism~\cite{arevalo2017gated}:
\begin{equation}
	\mathbf{z}^{\prime}=(1-q)\mathbf{z}+q\tilde{\mathbf{z}},
\end{equation}
where $\mathbf{z}^{\prime}$ denotes the final calibrated embedding. This fusion adaptively balances the original embedding and the modulated embedding, so that noisy frequency cues do not directly dominate the final representation.

To further suppress excessive modulation, we introduce a regularization term on the modulation parameters:
\begin{equation}
	\mathcal{L}_{ap}
	=
	\mathbb{E}\big[(\gamma-1)^2\big]
	+
	\mathbb{E}\big[\beta^2\big].
\end{equation}
where $\mathbb{E}(\cdot)$ denotes the expectation operator, which computes the average value of the enclosed quantity over samples. Therefore, the first term measures the average deviation of $\gamma$ from the identity value $1$, while the second term measures the average magnitude of $\beta$. This regularizer keeps the scaling factor close to the identity mapping and constrains the bias magnitude, thereby encouraging the modulation branch to perform only necessary sample-specific corrections.

\subsection{Batch-Wise Standardization on Brain Embeddings}
\label{sec:bn_whiten}
Although frequency-based modulation refines sample-specific embeddings, the resulting neural representations may still show feature-scale inconsistency and unstable batch-wise statistics. We apply batch-wise standardization before cross-modal similarity computation to stabilize the neural embedding distribution~\cite{ioffe2015batch}. Given a batch of brain embeddings $\mathbf{Z}\in\mathbb{R}^{B\times D}$, where $B$ is the batch size and $D$ is the embedding dimension, we compute the batch statistics for each feature dimension as:
\begin{equation}
	\mu_k = \frac{1}{B}\sum_{i=1}^{B} z_{ik},
	\sigma_k^2 = \frac{1}{B}\sum_{i=1}^{B}\left(z_{ik}-\mu_k\right)^2.
\end{equation}
Using these statistics, each embedding dimension is standardized as
\begin{equation}
	\tilde{z}_{ik}=\frac{z_{ik}-\mu_k}{\sqrt{\sigma_k^2+\epsilon}},
	\quad i=1,\ldots,B,; k=1,\ldots,D,
	\label{eq:bws}
\end{equation}
where $\epsilon>0$ is a small constant for numerical stability, and $\tilde{\mathbf{Z}}\in\mathbb{R}^{B\times D}$ denotes the standardized embedding matrix. This operation reduces scale mismatch across embedding dimensions and improves the stability of subsequent alignment learning.

\subsection{Low-Dimensional Adapter and Correlation Regularization}
After batch-wise stabilization of neural embeddings, we project both modalities into a compact shared subspace through a low-dimensional adapter, where bottleneck components are encouraged to capture transferable cross-modal structure~\cite{andrew2013deep,zbontar2021barlow}. Given an adapted feature $\tilde{\mathbf{z}}^{m}\in\mathbb{R}^{d}$, where $m\in\{e,v\}$ denotes the neural or visual modality, we first obtain the bottleneck representation
\[
\mathbf{h}^{m}=\mathbf{W}_{down}^{m}\tilde{\mathbf{z}}^{m}\in\mathbb{R}^{r},
\]
where $\mathbf{W}_{down}^{m}\in\mathbb{R}^{r\times d}$ is the down-projection matrix and $r\ll d$ is the bottleneck dimension. The refined feature is then reconstructed as
\begin{equation}
	\begin{split}
		\mathbf{z}_{out}^{m}
		=
		\mathrm{LayerNorm}\Big(
		\tilde{\mathbf{z}}^{m}
		+
		\mathbf{W}_{up}^{m}\big(
		\mathrm{Dropout}(
		\mathbf{h}^{m})
		\big)
		\Big), \\
		m\in\{e,v\},
	\end{split}
\end{equation}
where $\mathbf{W}_{up}^{m}\in\mathbb{R}^{d\times r}$ is the up-projection matrix, and $\mathbf{z}_{out}^{m}\in\mathbb{R}^{d}$ denotes the refined feature.

\subsubsection*{Correlation regularization in the low-dimensional space}

To further enforce structural consistency across modalities, we regularize the bottleneck representations produced by the down projection. For a mini-batch of size $B$, let $\mathbf{H}^{e}\in\mathbb{R}^{B\times r}$ and $\mathbf{H}^{v}\in\mathbb{R}^{B\times r}$ denote the stacked neural and visual bottleneck features, respectively. We then introduce a correlation regularization term on $\mathbf{H}^{e}$ and $\mathbf{H}^{v}$ to encourage aligned and non-redundant structure in the shared low-dimensional subspace~\cite{hotelling1992relations}.

We first standardize each feature dimension across the batch:
\begin{equation}
	\bar{\mathbf{H}}^{m}
	=
	\frac{\mathbf{H}^{m}-\mu^{m}}{\sigma^{m}+\epsilon},
	\qquad m\in\{e,v\},
\end{equation}
where $\mu^{m}$ denotes the mean of each feature dimension computed over the batch for modality $m$, and $\sigma^{m}$ denotes the corresponding standard deviation. The constant $\epsilon$ is a small positive value added to avoid numerical instability during division.

Based on the standardized features, we compute the cross-modal correlation matrix:
\begin{equation}
	\mathbf{C}
	=
	\frac{(\bar{\mathbf{H}}^{e})^{\top}\bar{\mathbf{H}}^{v}}{B},
	\qquad
	\mathbf{C}\in\mathbb{R}^{r\times r}.
\end{equation}
where $C \in \mathbb{R}^{r \times r}$ denotes the cross-modal correlation matrix, and the division by $B$ normalizes the correlation by the batch size.

The cross-correlation regularization is defined as
\begin{equation}
	\mathcal{L}_{\mathrm{corr}}
	=
	\sum_{i=1}^{r}(1-C_{ii})^{2}
	+
	\lambda_{\mathrm{off}}
	\sum_{i=1}^{r}\sum_{\substack{j=1\\ j\neq i}}^{r} C_{ij}^{2},
\end{equation}
where the diagonal terms $C_{ii}$ measure the correlation between corresponding dimensions of the two modalities. Pushing $C_{ii}$ toward 1 encourages the aligned components to be highly correlated. The off diagonal terms $C_{ij}$, where $i \neq j$, capture inter-dimension correlations. Suppressing these terms reduces redundancy across dimensions and promotes more decorrelated representations. The coefficient $\lambda_{\mathrm{off}}$ balances the off diagonal penalty.
\begin{table*}[t]
	\caption{Top-1 and Top-5 accuracy (\%) for 200-way zero-shot retrieval on THINGS-EEG. The table reports results in both the intra-subject setting, where training and testing are performed on the same subject, and the inter-subject setting, where one subject is held out for testing.}
	\label{tab:things_eeg_200way}
	\centering
	\footnotesize
	\setlength{\tabcolsep}{3.2pt}
	\renewcommand{\arraystretch}{1.10}
	
	\resizebox{\textwidth}{!}{%
		\begin{tabular}{l*{11}{cc}}
			\toprule
			\multirow{2}{*}{Method}
			& \multicolumn{2}{c}{Subj.\ 1}
			& \multicolumn{2}{c}{Subj.\ 2}
			& \multicolumn{2}{c}{Subj.\ 3}
			& \multicolumn{2}{c}{Subj.\ 4}
			& \multicolumn{2}{c}{Subj.\ 5}
			& \multicolumn{2}{c}{Subj.\ 6}
			& \multicolumn{2}{c}{Subj.\ 7}
			& \multicolumn{2}{c}{Subj.\ 8}
			& \multicolumn{2}{c}{Subj.\ 9}
			& \multicolumn{2}{c}{Subj.\ 10}
			& \multicolumn{2}{c}{Avg} \\
			\cmidrule(lr){2-3}\cmidrule(lr){4-5}\cmidrule(lr){6-7}\cmidrule(lr){8-9}\cmidrule(lr){10-11}
			\cmidrule(lr){12-13}\cmidrule(lr){14-15}\cmidrule(lr){16-17}\cmidrule(lr){18-19}\cmidrule(lr){20-21}\cmidrule(lr){22-23}
			& Top-1 & Top-5
			& Top-1 & Top-5
			& Top-1 & Top-5
			& Top-1 & Top-5
			& Top-1 & Top-5
			& Top-1 & Top-5
			& Top-1 & Top-5
			& Top-1 & Top-5
			& Top-1 & Top-5
			& Top-1 & Top-5
			& Top-1 & Top-5 \\
			\midrule
			
			\multicolumn{23}{c}{\textbf{Intra-subject:} train and test on one subject} \\
			\midrule
			BraVL~\cite{du2023decoding}   &  6.1 & 17.9 &  4.9 & 14.9 &  5.6 & 17.4 &  5.0 & 15.1 &  4.0 & 13.4 &  6.0 & 18.2 &  6.5 & 20.4 &  8.8 & 23.7 &  4.3 & 14.0 &  7.0 & 19.7 &  5.8 & 17.5 \\
			NICE~\cite{nice60}    & 13.2 & 39.5 & 13.5 & 40.3 & 14.5 & 42.7 & 20.6 & 52.7 & 10.1 & 31.5 & 16.5 & 44.0 & 17.0 & 42.1 & 22.9 & 56.1 & 15.4 & 41.6 & 17.4 & 45.8 & 16.1 & 43.6 \\
			NICE-SA~\cite{nice60} & 13.3 & 40.2 & 12.1 & 36.1 & 15.3 & 39.6 & 15.9 & 49.0 &  9.8 & 34.4 & 14.2 & 42.4 & 17.9 & 43.6 & 18.2 & 50.2 & 14.4 & 38.7 & 16.0 & 42.8 & 14.7 & 41.7 \\
			NICE-GA~\cite{nice60} & 15.2 & 40.1 & 13.9 & 40.1 & 14.7 & 42.7 & 17.6 & 48.9 &  9.0 & 29.7 & 16.4 & 44.4 & 14.9 & 43.1 & 20.3 & 52.1 & 14.1 & 39.7 & 19.6 & 46.7 & 15.6 & 42.8 \\
			ATM-S~\cite{li2024visual}    & 25.6 & 60.4 & 22.0 & 54.5 & 25.0 & 62.4 & 31.4 & 60.9 & 12.9 & 43.0 & 21.3 & 51.1 & 30.5 & 61.5 & 38.8 & 72.0 & 34.4 & 51.5 & 29.1 & 63.5 & 28.5 & 60.4 \\
			VE-SDN~\cite{vesdn10} & 32.6 & 63.7 & 34.4 & 69.9 & 38.7 & 73.5 & 39.8 & 72.0 & 29.4 & 58.6 & 34.5 & 68.8 & 34.5 & 68.3 & 49.3 & 79.8 & 39.0 & 69.6 & 39.8 & 75.3 & 37.2 & 69.9 \\
			UBP~\cite{wu2025bridging}   & 41.2 & 70.5 & 51.2 & 80.9 & 51.2 & 82.0 & 51.1 & 76.9 & 42.2 & 72.8 & 57.5 & 83.5 & 49.0 & 79.9 & 58.6 & 85.8 & 45.1 & 76.2 & 61.5 & 88.2 & 50.9 & 79.7 \\
			\mcell{Ours}   & \mcell{61.0} & \mcell{85.5} & \mcell{61.0} & \mcell{86.0} & \mcell{65.0} & \mcell{87.0} & \mcell{60.5} & \mcell{87.5} & \mcell{54.5} & \mcell{80.0} & \mcell{69.0} & \mcell{86.0} & \mcell{64.0} & \mcell{87.5} & \mcell{69.0} & \mcell{91.5} & \mcell{59.0} & \mcell{86.5} & \mcell{71.0} & \mcell{91.0} & \mcell{63.40} & \mcell{86.85} \\
			\midrule
			
			\multicolumn{23}{c}{\textbf{Inter-subject:} leave one subject out for test} \\
			\midrule
			BraVL    &  2.3 &  8.0 &  1.5 &  6.3 &  1.4 &  5.9 &  1.7 &  6.7 &  1.5 &  5.6 &  1.8 &  7.2 &  2.1 &  8.1 &  2.2 &  7.6 &  1.6 &  6.4 &  2.3 &  8.5 &  1.8 &  7.0 \\
			NICE    &  7.6 & 22.8 &  5.9 & 20.5 &  6.0 & 22.3 &  6.3 & 20.7 &  4.4 & 18.3 &  5.6 & 22.2 &  5.6 & 19.7 &  6.3 & 22.0 &  5.7 & 17.6 &  8.4 & 28.3 &  6.2 & 21.4 \\
			NICE-SA  &  7.0 & 22.6 &  6.6 & 23.2 &  7.5 & 23.7 &  5.4 & 21.4 &  6.4 & 22.2 &  7.5 & 22.5 &  3.8 & 19.1 &  8.5 & 24.4 &  7.4 & 22.3 &  9.8 & 29.6 &  7.0 & 23.1 \\
			NICE-GA  &  5.9 & 21.4 &  6.4 & 22.7 &  5.5 & 20.1 &  6.1 & 21.0 &  4.7 & 19.5 &  6.2 & 22.5 &  5.9 & 19.1 &  7.3 & 25.3 &  4.8 & 18.3 &  6.2 & 26.3 &  5.9 & 21.6 \\
			ATM-S     & 10.5 & 26.8 &  7.1 & 24.8 & 11.9 & 33.8 & 14.7 & 39.4 &  7.0 & 23.9 & 11.1 & 35.8 & 16.1 & 43.5 & 15.0 & 40.3 &  4.9 & 22.7 & 20.5 & 46.5 & 11.8 & 33.7 \\
			UBP    & 11.5 & 29.7 & 15.5 & 40.0 &  9.8 & 27.0 & 13.0 & 32.3 &  8.8 & 33.8 & 11.7 & 31.0 & 10.2 & 23.8 & 12.2 & 32.2 & 15.5 & 40.5 & 16.0 & 43.5 & 12.4 & 33.4 \\
			\mcell{Ours}    & \mcell{28.0} & \mcell{61.5} & \mcell{23.5} & \mcell{57.0} & \mcell{21.0} & \mcell{49.0} &\mcell{21.0} & \mcell{49.0} & \mcell{19.5} & \mcell{47.0} & \mcell{26.5} & \mcell{57.5} & \mcell{25.5} & \mcell{48.5} & \mcell{18.0} & \mcell{44.0} & \mcell{26.0} & \mcell{56.5} & \mcell{31.5} & \mcell{64.5} & \mcell{24.05} & \mcell{53.45} \\
			\bottomrule
		\end{tabular}%
	}
\end{table*}

\begin{table}[t]
	\caption{Top-1 and Top-5 accuracy (\%) for 200-way zero-shot retrieval on THINGS-MEG. The table reports results in both the intra-subject setting and the inter-subject setting with leave one subject out evaluation.}
	\label{tab:things_meg_200way}
	\centering
	\footnotesize
	\setlength{\tabcolsep}{3.5pt}
	\renewcommand{\arraystretch}{1.10}
	
	\resizebox{\columnwidth}{!}{%
		\begin{tabular}{l*{5}{cc}}
			\toprule
			\multirow{2}{*}{Method}
			& \multicolumn{2}{c}{Subj.\ 1}
			& \multicolumn{2}{c}{Subj.\ 2}
			& \multicolumn{2}{c}{Subj.\ 3}
			& \multicolumn{2}{c}{Subj.\ 4}
			& \multicolumn{2}{c}{Avg} \\
			\cmidrule(lr){2-3}\cmidrule(lr){4-5}\cmidrule(lr){6-7}\cmidrule(lr){8-9}\cmidrule(lr){10-11}
			& Top-1 & Top-5
			& Top-1 & Top-5
			& Top-1 & Top-5
			& Top-1 & Top-5
			& Top-1 & Top-5 \\
			\midrule
			
			\multicolumn{11}{c}{\textbf{Intra-subject:} train and test on one subject} \\
			\midrule
			NICE     &  9.6 & 27.8 & 18.5 & 47.8 & 14.2 & 41.6 &  9.0 & 26.6 & 12.8 & 36.0 \\
			NICE-SA  &  9.8 & 27.8 & 18.6 & 46.4 & 10.5 & 38.4 & 11.7 & 27.2 & 12.7 & 35.0 \\
			NICE-GA  &  8.7 & 30.5 & 21.8 & 56.6 & 16.5 & 49.7 & 10.3 & 32.3 & 14.3 & 42.3 \\
			UBP      & 15.0 & 38.0 & 46.0 & 80.5 & 27.3 & 59.0 & {18.5} & 43.5 & 26.7 & 55.2 \\
			Ours & \textbf{22.0} & \textbf{52.5} & \textbf{63.5} & \textbf{89.0} & \textbf{43.9} & \textbf{74.5} & \textbf{28.5} & \textbf{50.0} & \textbf{39.25} & \textbf{66.50} \\
			\midrule
			
			\multicolumn{11}{c}{\textbf{Inter-subject:} leave one subject out for test} \\
			\midrule
			UBP      &  2.0 &  5.7 &  1.5 & 17.2 &  2.7 & 10.5 &  2.5 &  8.0 &  2.2 & 10.4 \\
			Ours &  \textbf{3.5} &  \textbf{9.0} &  \textbf{7.5} & \textbf{25.0} & \textbf{4.0} & \textbf{19.0} &  \textbf{4.5} &  \textbf{17.0} &  \textbf{4.875} & \textbf{17.50} \\
			\bottomrule
		\end{tabular}%
	}
\end{table}
\subsection{Overall Loss}
\label{sec:overall}
\subsubsection*{Phase-Stability-Derived Control Signal}
We compute a sample-wise control score from phase-circular variance and use it to modulate the contrastive temperature. This mechanism adjusts optimization strength according to phase stability, rather than serving as an explicit estimator of sample difficulty. To this end, we define a sample-wise control score based on phase-circular variance:
\begin{equation}
	r_i = 1 - \frac{1}{C}\sum_{c=1}^{C} v_{i,c}^{\phi},
\end{equation}
where $r_i$ denotes the control score of the $i$-th sample, $v_{i,c}^{\phi}$ is the phase-circular variance of the $c$-th channel for the $i$-th sample, and $C$ is the number of neural channels. A larger $r_i$ indicates stronger phase stability and is used only to modulate optimization strength during contrastive learning.
\subsubsection*{Adaptive temperature mapping}
We then map the control score to a sample-specific contrastive temperature~\cite{chen2020simple}:
\begin{equation}
	\tau_i = \mathrm{softplus}(a - b r_i) + \tau_{\min},
\end{equation}
where $\tau_i$ is the adaptive temperature of the $i$-th sample, $a$ and $b$ are learnable parameters, and $\tau_{\min}$ is the minimum temperature introduced for numerical stability.

Let $\hat{\mathbf{z}}_i^{e}$ and $\hat{\mathbf{z}}j^{v}$ denote the normalized EEG and image embeddings of the $i$th and $j$-th samples, respectively. Their cross-modal similarity is defined as
\begin{equation}
	s_{ij} = (\hat{\mathbf{z}}_i^{e})^{\top}\hat{\mathbf{z}}j^{v},
\end{equation}
where $s{ij}$ is the similarity between the $i$th EEG embedding and the $j$th image embedding.
\subsubsection*{Overall training objective}
Based on the adaptive temperature and the pairwise similarity, we define the bidirectional sample-wise contrastive losses and the overall adaptive symmetric contrastive loss jointly as:
\begin{equation}
	\begin{split}
		\mathcal{L}_{\mathrm{adapt}}
		&=
		\frac{1}{2B}
		\sum_{i=1}^{B}
		\Bigg[
		-\log
		\frac{\exp\left(s_{ii}/\tau_i\right)}
		{\sum_{j=1}^{B}\exp\left(s_{ij}/\tau_i\right)}
		\\
		&\qquad
		-
		\log
		\frac{\exp\left(s_{ii}/\tau_i\right)}
		{\sum_{j=1}^{B}\exp\left(s_{ji}/\tau_i\right)}
		\Bigg].
	\end{split}
\end{equation}
where $B$ is the mini-batch size, $s_{ii}$ is the similarity between the $i$th EEG sample and its matched image, and $\tau_i$ is the adaptive temperature assigned to the $i$-th sample according to its control score. The first term inside the brackets corresponds to the EEG to image contrastive direction, and the second term corresponds to the image to EEG contrastive direction. We use the same sample-specific temperature in both contrastive directions so that the reliability-conditioned sharpening remains consistent for each neural anchor pair.

Finally, together with the amplitude and phase modulation regularization $\mathcal{L}{ap}$ and the low-dimensional correlation regularization $\mathcal{L}{corr}$ defined above, the overall training objective is written as
\begin{equation}
	\mathcal{L}_{\mathrm{total}}
	=
	\mathcal{L}_{\mathrm{adapt}}
	+
	\lambda_{corr}\mathcal{L}_{corr}
	+
	\lambda_{ap}\mathcal{L}_{ap}.
\end{equation}
where $\lambda{corr}$ and $\lambda_{ap}$ are weighting coefficients for the low-dimensional correlation regularization and the amplitude and phase modulation regularization, respectively. In our implementation, we set $\lambda_{corr}=0.05$ and $\lambda_{ap}=0.1$.

\label{gen_inst}
\section{Experiments and Results}
\label{headings}
Section~\ref{headings} presents the experimental protocols and the main retrieval results. We first describe the EEG and MEG datasets together with the implementation settings, including optimization details and the visual and neural backbones. We then report the quantitative retrieval performance, followed by qualitative visualization and diagnostic analyses to further interpret the behavior of the proposed framework.
\subsection{Dataset}
\paragraph{EEG dataset}

THINGS-EEG~\cite{gifford2022eeg} contains recordings from 10 subjects. The training split covers 1,654 object concepts, with 10 images for each concept, and each image is repeated 4 times per subject. The test split contains 200 concepts, with 1 image for each concept, and each test image is repeated 80 times per subject. For preprocessing, we follow the pipeline in~\cite{gifford2022eeg}. To improve the signal-to-noise ratio, repeated trials elicited by the same stimulus are averaged, resulting in 16,540 training samples and 200 test samples for each subject.

\paragraph{MEG dataset}

THINGS-MEG~\cite{hebart2023thingsdata} includes recordings from 4 participants acquired with 271 MEG channels. The training split consists of 1,854 concepts, 12 images for each concept, and 1 repetition for each image, whereas the test split consists of 200 concepts, 1 image for each concept, and 12 repetitions for each image. We adopt the same experimental setting and preprocessing protocol as~\cite{hebart2023thingsdata}. 
\subsection{Experimented Setup}
\paragraph{Environment And Implementation Details}
All experiments were implemented in Python 3.8.19 with CUDA 12.0 and PyTorch 2.4.1. Additional dependencies are provided in \texttt{requirements.txt}. Training was conducted on a workstation equipped with six NVIDIA Tesla T4 GPUs, each with 14.5GB of memory. We trained all models for 50 epochs with a batch size of $B=1024$. Optimization was performed using AdamW~\cite{frank2019decoupled} with a weight decay of $10^{-4}$. Following the standard protocol, the learning rate was set to $10^{-4}$ for the intra-subject setting and to $10^{-5}$ for the inter-subject setting. Early stopping was performed based on validation performance to mitigate overfitting.

\begin{figure*}[t]
	\centering
	\includegraphics[width=\textwidth, trim=2mm 2mm 2mm 2mm, clip]{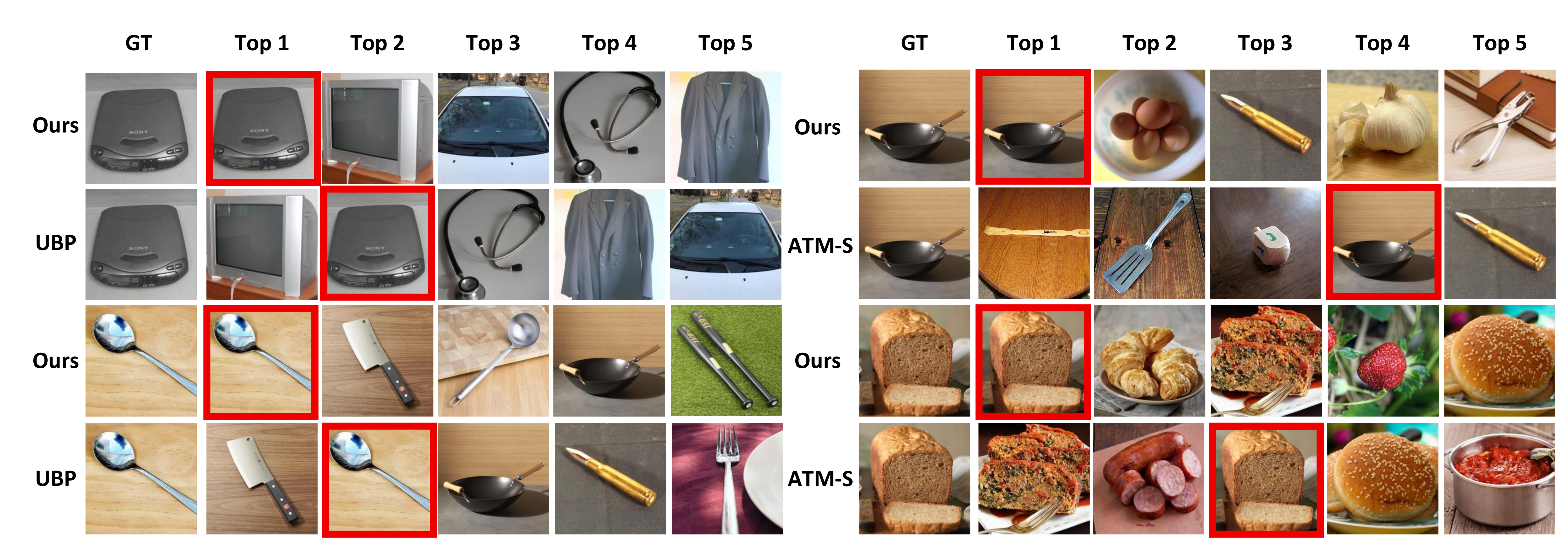}
	\caption{Qualitative comparison of neural-to-image retrieval results. The first column shows the ground-truth target image (GT), and the remaining columns show the Top-1 to Top-5 retrieved results for each method. Red boxes denote correct matches.}
	\label{fig:top-5-retrival}
\end{figure*}

\paragraph{Vision Encoders}

For the visual branch, we use pretrained CLIP image encoders from OpenCLIP~\cite{gadre2023datacomp, cherti2023reproducible, ilharco2021openclip}. These encoders are trained on large-scale image-text data collected from diverse sources under different training configurations, and they provide general-purpose visual representations. In our experiments, we evaluate multiple visual backbones, including RN50, RN101, ViT-B-16, ViT-B-32, ViT-L-14, ViT-H-14, and ViT-g-14. Unless otherwise specified, RN50 is used as the default visual encoder.

\paragraph{Brain Encoders}

In the main model, neural signals are encoded by the brain encoder described in the Method section. This encoder serves as the default neural backbone in our framework and is designed for efficient cross-modal alignment in the shared semantic space. To further evaluate the robustness and generalizability of the proposed method, we also consider several alternative EEG backbones, including ShallowNet~\cite{schirrmeister2017deep}, DeepNet~\cite{schirrmeister2017deep}, EEGProject~\cite{wu2025bridging}, EEGNet~\cite{lawhern2018eegnet}, and TSConv~\cite{song2023decoding}. For all these variants, we append the same projection head to the backbone output so that the comparison focuses on the encoder itself rather than on differences in downstream alignment design.

\subsection{Comparison with Baselines}
Tables~\ref{tab:things_eeg_200way} and ~\ref{tab:things_meg_200way} report the main quantitative retrieval results on THINGS-EEG and THINGS-MEG under both intra-subject and inter-subject evaluation protocols. We compare the proposed framework with representative baselines to assess not only within-subject retrieval accuracy but also cross-subject generalization under heterogeneous neural conditions.

\subsection{Semantic Alignment Analysis and Visualization}
\begin{figure}[t]
	\centering
	\includegraphics[width=\columnwidth,trim=10mm 2mm 2mm 2mm, clip]{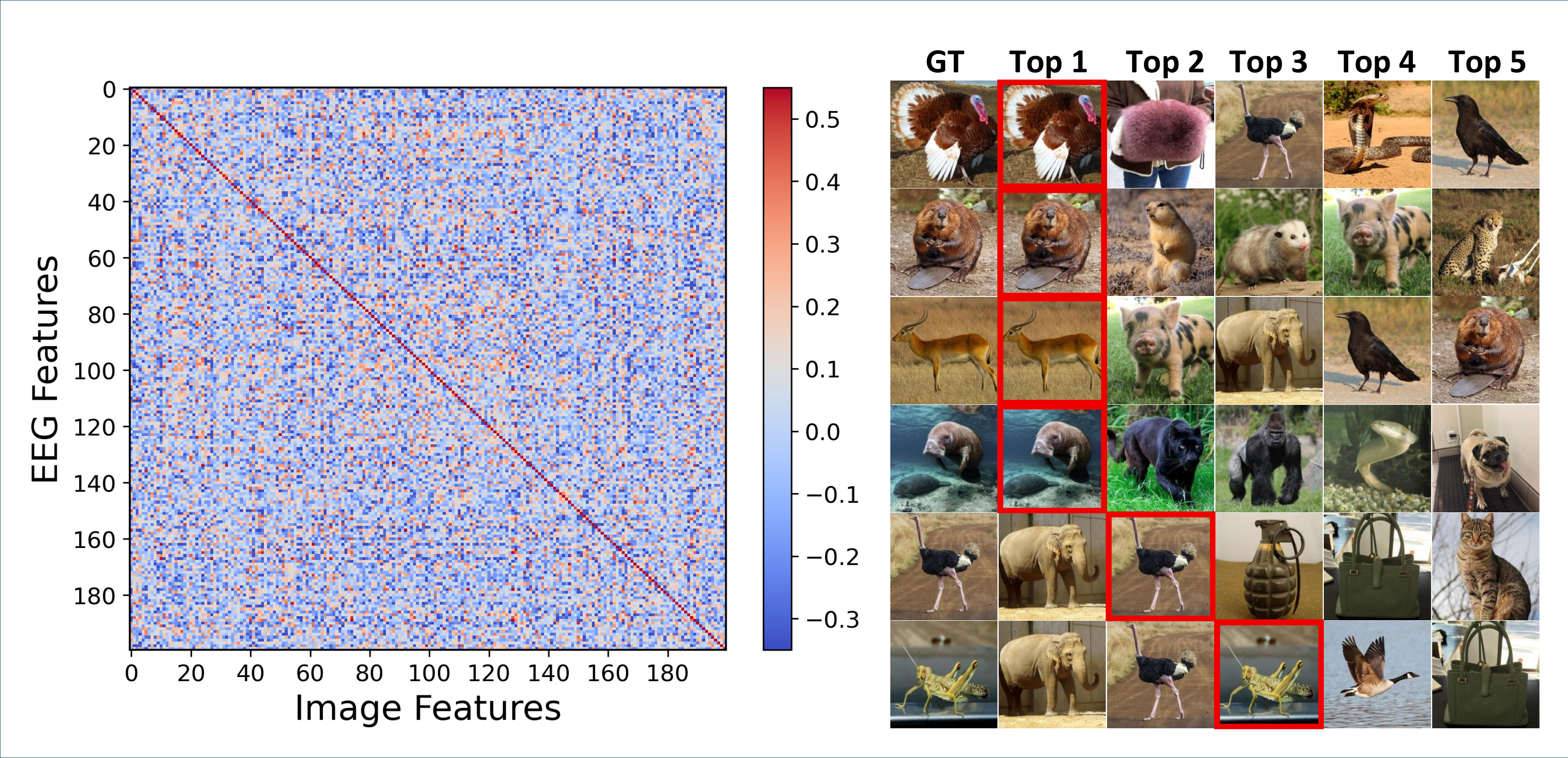}
	\caption{Visualization of EEG--image cross-modal alignment. Left: correlation heatmap between EEG features and image features in the shared embedding space. Right: retrieval examples, where red boxes denote the ground-truth target images.}
	\label{fig:retrieval_vis}
\end{figure}

As shown in Fig.~\ref{fig:retrieval_vis}, matched EEG-image pairs appear more concentrated in the shared embedding space after training. The retrieval examples also suggest that the learned representation captures coarse semantic similarity reasonably well. However, several failure cases still occur among semantically related categories, indicating that fine-grained discrimination remains challenging.

\subsection{Synergistic Effect of Frequency Priors and Adaptive Supervision}

\begin{figure}[t]
	\centering
	\includegraphics[width=0.45\textwidth]{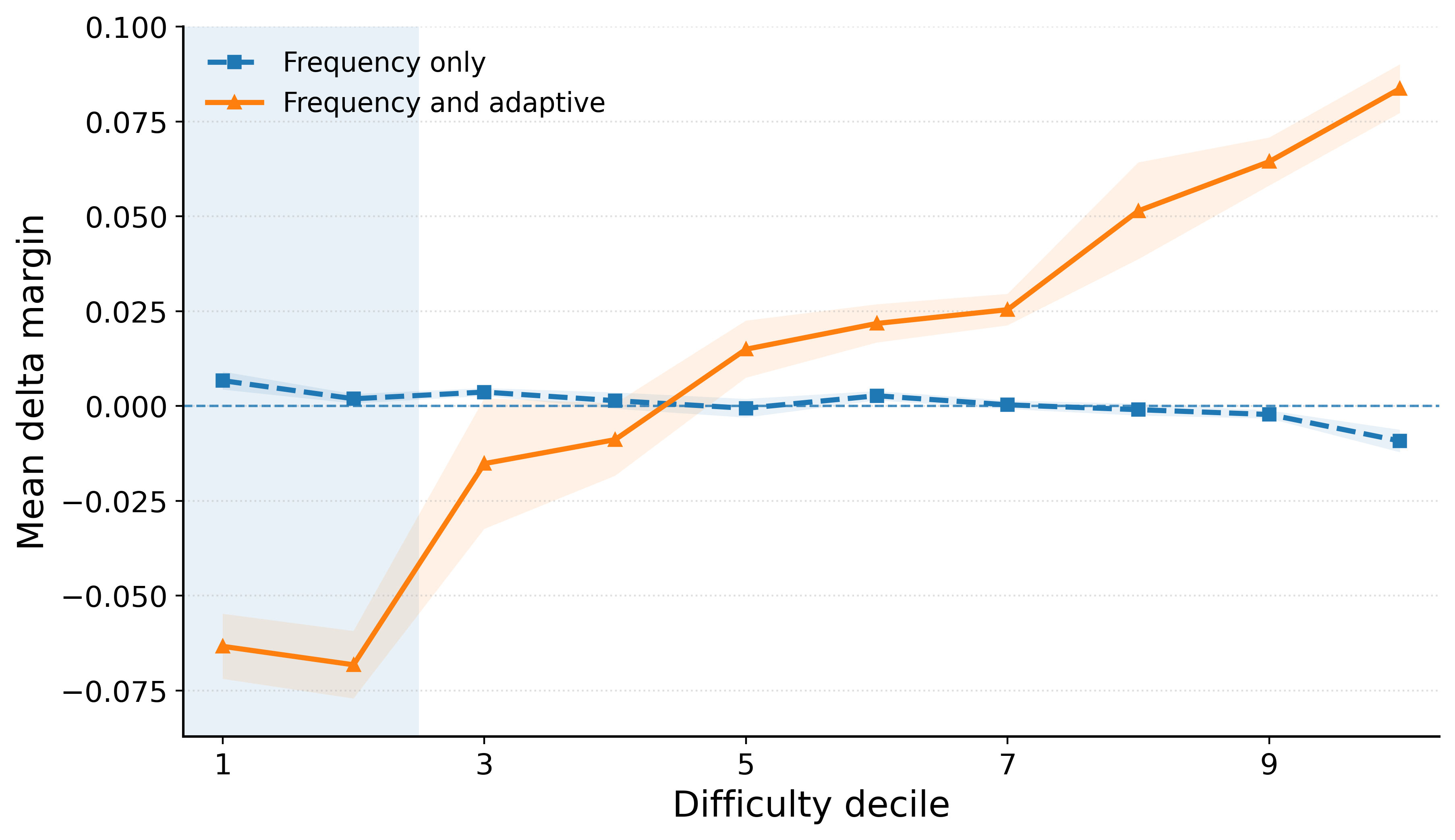}
	\caption{Gain trends of the \textit{frequency-only} and joint \textit{frequency-and-adaptive} variants across bins defined by the baseline retrieval margin. Lower baseline margin indicates more challenging samples. The \textit{frequency-only} variant remains near zero across most bins, whereas the joint variant shows larger gains in the lower-margin region. }
	\label{fig:joint_effect_difficulty}
\end{figure}

Fig.~\ref{fig:joint_effect_difficulty} compares the gain trends of the \emph{frequency-only} and \emph{frequency-and-adaptive} variants across bins defined by the baseline retrieval margin, while Table~III summarizes their average gains over five random seeds. The \emph{frequency-only} variant remains close to zero across most bins and produces only marginal changes in retrieval margin, retrieval rank, and Top-1 accuracy, indicating limited standalone benefit. In contrast, the joint \emph{frequency-and-adaptive} design yields clearer gains, especially in the lower-margin region, where the baseline samples are more challenging. This trend is broadly consistent with Table~III: the joint design improves the overall $\Delta$Margin, $\Delta$Rank, and $\Delta$Top-1, whereas the \emph{frequency-only} variant remains nearly neutral. Within the hardest twenty percent subset, the most evident gains of the joint design appear in $\Delta$Rank and $\Delta$Top-1, while the change in $\Delta$Margin is less consistent. These results suggest that frequency priors are more effective when combined with adaptive supervision than when used alone.
\begin{table*}[t]
	\centering
	\scriptsize
	\setlength{\tabcolsep}{3pt}
	\renewcommand{\arraystretch}{1.08}
	\caption{Summary of the effects of adaptive supervision, frequency modulation, and their joint design relative to the baseline over five random seeds. Each entry is reported as mean $\pm$ standard deviation. Positive $\Delta$Rank indicates that the compared method achieves a better, that is, smaller, retrieval rank than the baseline.}
	\label{tab:joint_effect_summary}
	\resizebox{\textwidth}{!}{%
		\begin{tabular}{lccccccccc}
			\toprule
			\textbf{Comparison} & $\Delta$Margin & $\Delta$Rank & $\Delta$Top-1 
			& \textbf{Hardest20} $\Delta$Margin & \textbf{Hardest20} $\Delta$Rank & \textbf{Hardest20} $\Delta$Top-1
			& \textbf{Easiest20} $\Delta$Margin & \textbf{Easiest20} $\Delta$Rank & \textbf{Easiest20} $\Delta$Top-1 \\
			\midrule			
			Baseline vs.\ frequency-only
			& $0.0004 \pm 0.0021$ & $-0.064 \pm 0.177$ & $-0.003 \pm 0.010$
			& $0.0043 \pm 0.0028$ & $0.055 \pm 0.427$ & $0.005 \pm 0.011$
			& $-0.0057 \pm 0.0039$ & $0.000 \pm 0.000$ & $0.000 \pm 0.000$ \\
			
			Baseline vs.\ frequency-and-adaptive
			& $0.0106 \pm 0.0029$ & $0.464 \pm 0.452$ & $0.034 \pm 0.019$
			& $-0.0658 \pm 0.0098$ & $4.790 \pm 2.298$ & $0.130 \pm 0.021$
			& $0.0740 \pm 0.0104$ & $-0.075 \pm 0.098$ & $-0.030 \pm 0.033$ \\
			\bottomrule
		\end{tabular}%
	}
\end{table*}

\subsection{Control-Signal-Stratified Gain Analysis}

\begin{figure}[t]
	\centering
	\subfloat[Top-1 gain.]{
		\includegraphics[width=0.48\linewidth]{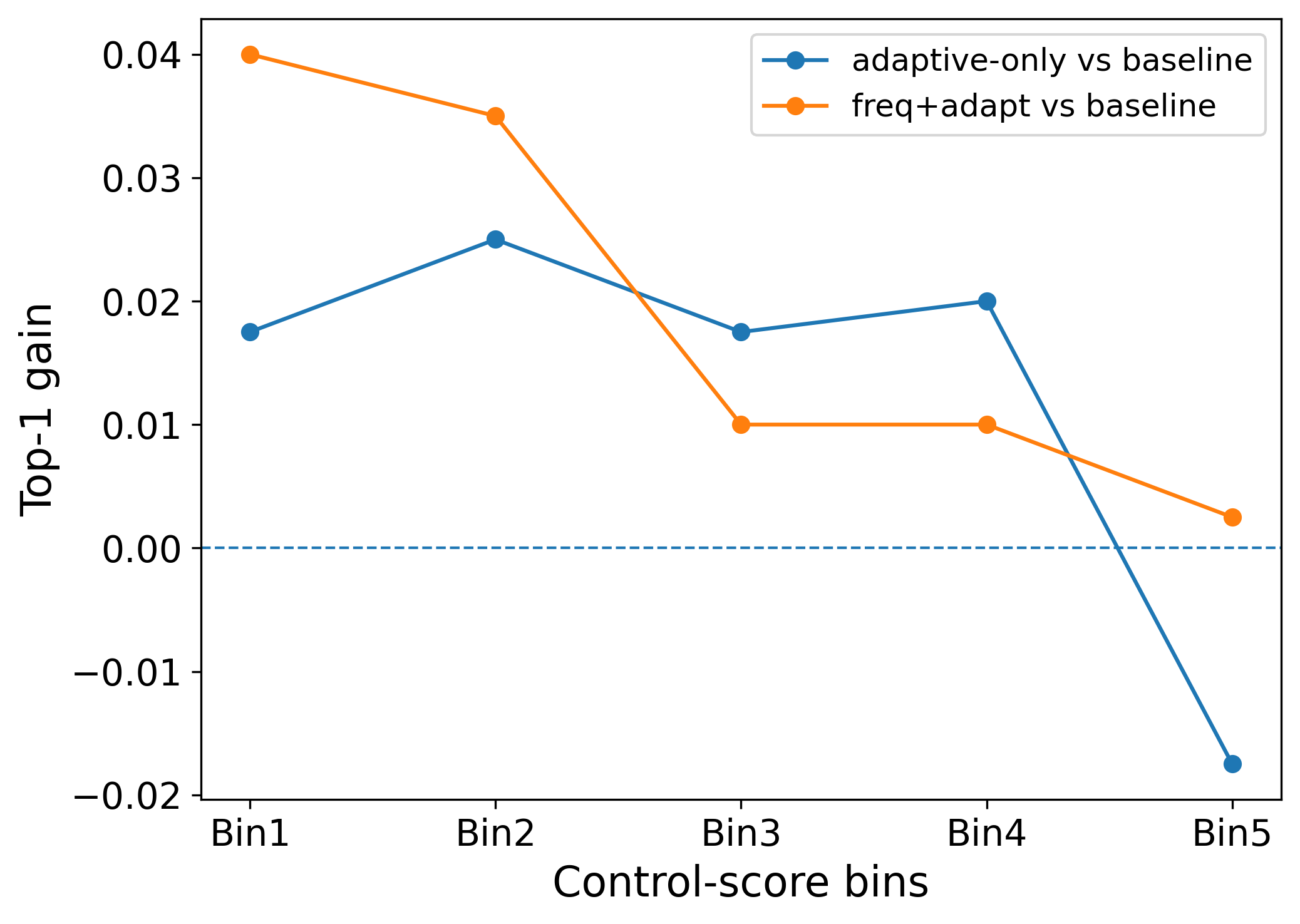}
		\label{fig:reliability_bin_top1_gain}
	}
	\hfill
	\subfloat[Margin gain.]{
		\includegraphics[width=0.48\linewidth]{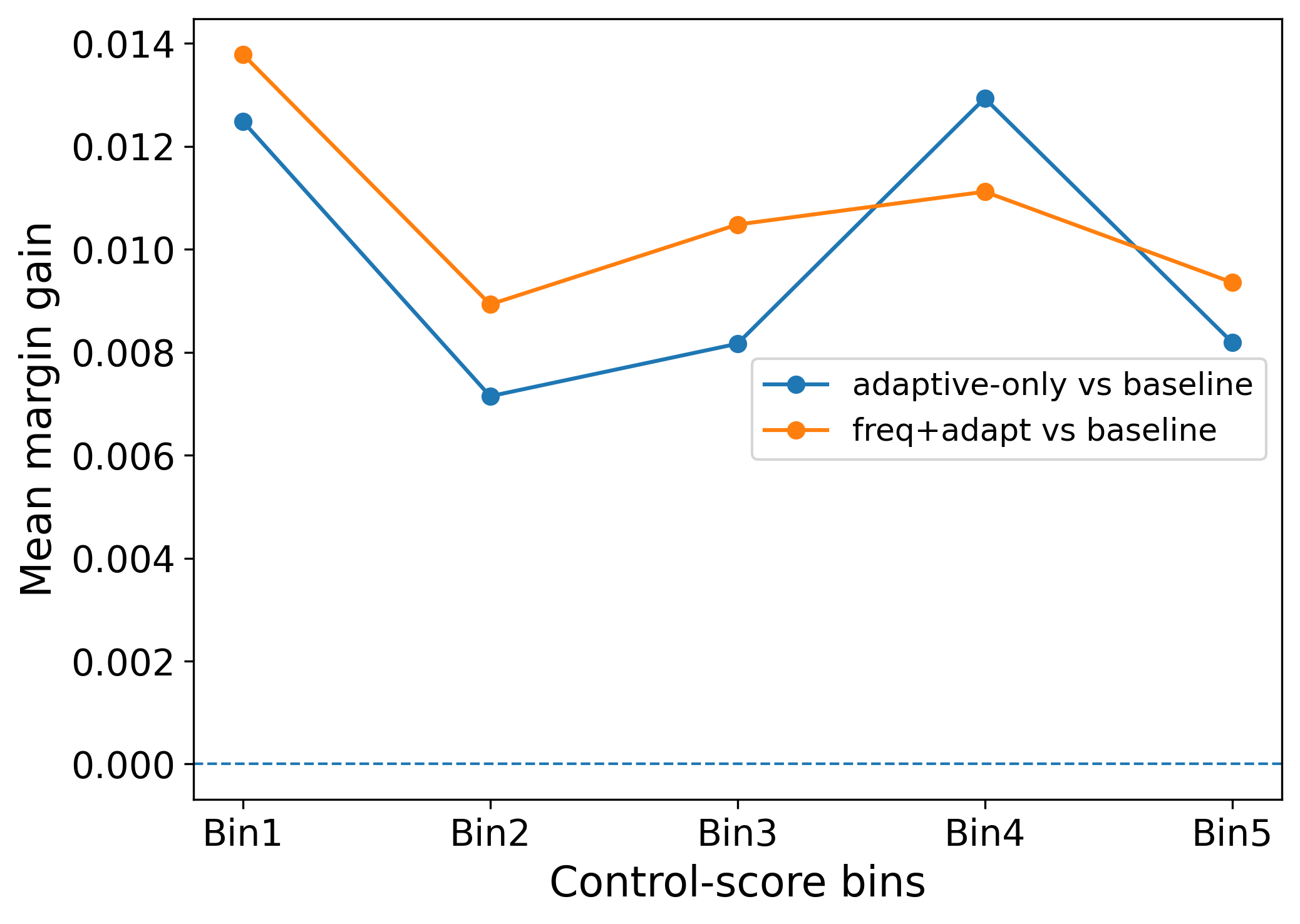}
		\label{fig:reliability_bin_margin_gain}
	}
	\caption{Gain analysis across five bins defined by the phase-derived control score estimated from the adaptive-only model. The bins reflect control-score grouping rather than ground-truth difficulty strata.}
	\label{fig:reliability_bin_gain}
\end{figure}

To examine how the gains vary across samples grouped by the phase-derived control signal, we partition the test samples into five bins according to the control scores produced by the \emph{adaptive-only} model. These bins are defined by the control signal. Therefore, the observed gains should be interpreted as differences in optimization responsiveness across control-score groups. We then compare the gains of \emph{adaptive-only} and \emph{frequency-and-adaptive} relative to the baseline. As shown in Fig.~\ref{fig:reliability_bin_gain}, the gains are not uniformly distributed across bins. Instead, the improvements are more pronounced in the lower-score bins, particularly for Top-1 accuracy and retrieval margin. The gains of the \emph{adaptive-only} variant become weaker as the control score increases and even turn negative in Top-1 accuracy in the highest-score bin. Moreover, the joint \emph{frequency-and-adaptive} design generally yields larger gains than \emph{adaptive-only} in the lowest-score bins. These results indicate that the gains of adaptive supervision are
more pronounced in lower-control-score groups and that frequency
modulation further strengthens this effect in those groups.

\subsection{Diagnostic Analysis of the Phase-Derived Control Signal}

\begin{figure}[t]
	\centering
	\subfloat[Pooled control signal versus margin.\label{fig:pooled_reliability_margin}]
	{\includegraphics[width=0.48\linewidth, trim=1 10 1 5,clip]{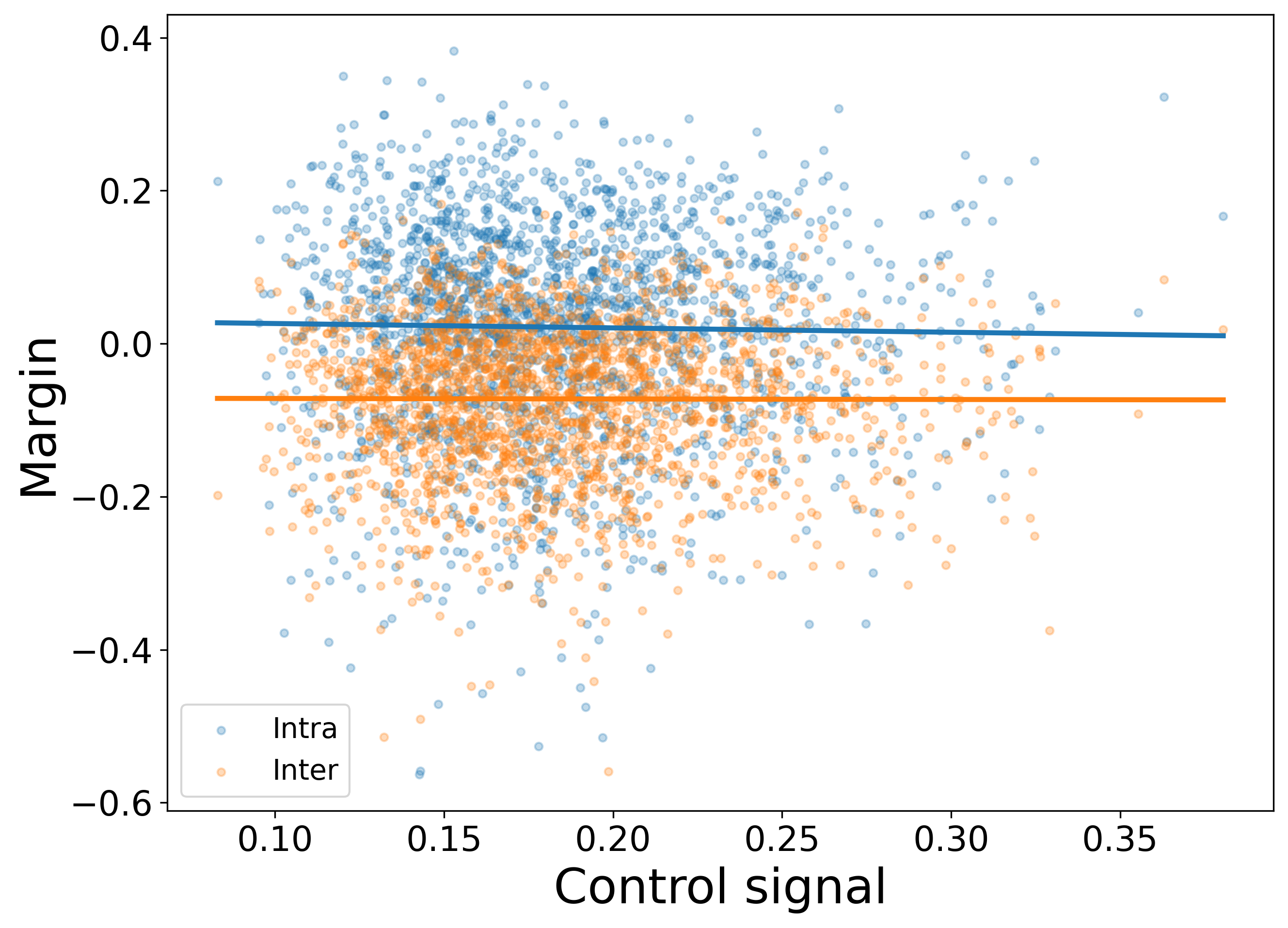}}
	\hfill
	\subfloat[Subject-wise Spearman correlations.\label{fig:subjectwise_spearman}]
	{\includegraphics[width=0.48\linewidth, trim=1 0 1 8,clip]{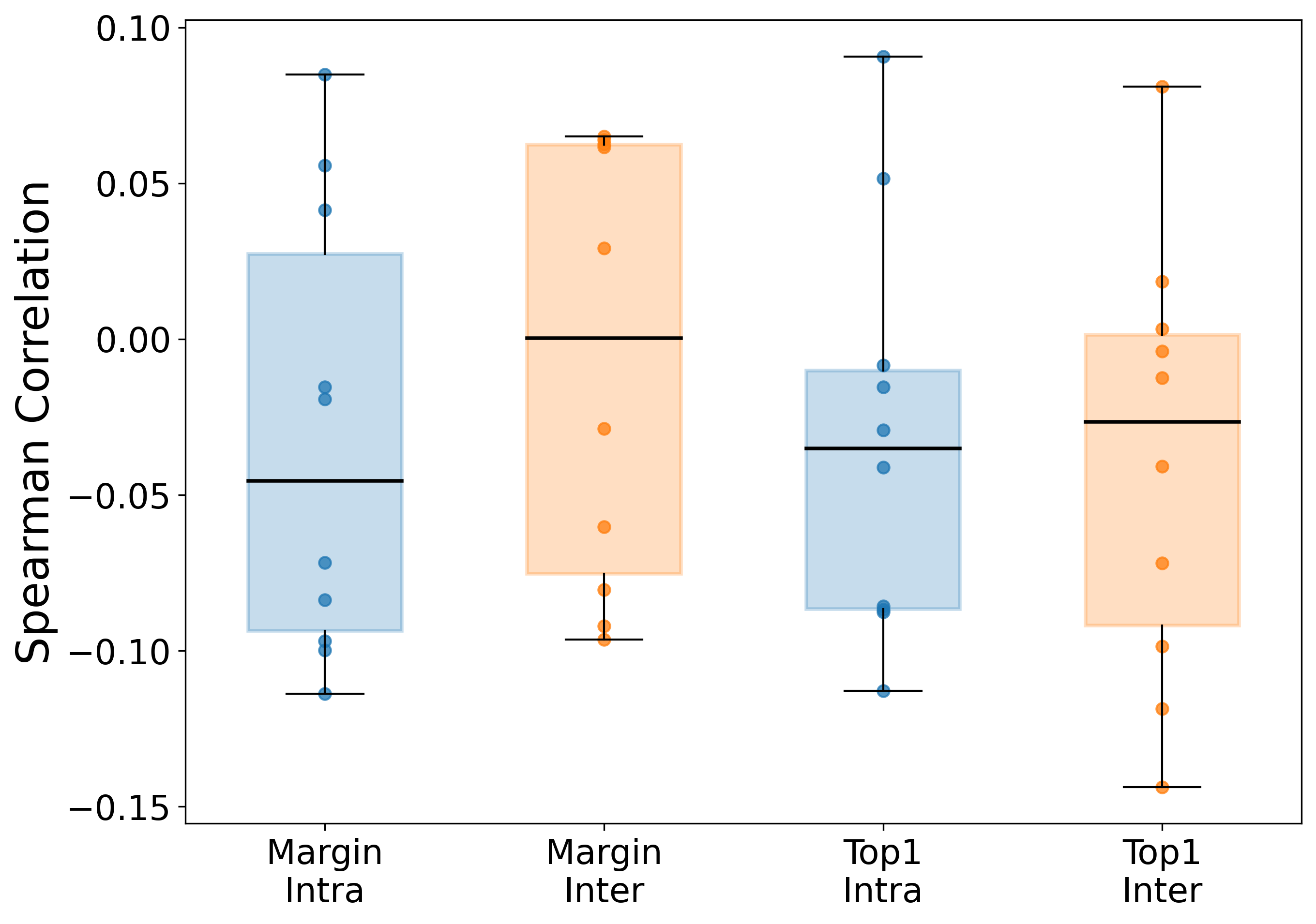}}
	\caption{Diagnostic visualization of the relationship between the phase-derived control signal and retrieval diagnostics. The pooled sample-level trend is nearly flat, and the subject-wise correlations remain centered near zero, suggesting that the current control signal is not strongly associated with sample-wise retrieval outcomes.}
	\label{fig:reliability_diagnostic}
\end{figure}

Fig.~\ref{fig:reliability_diagnostic} and Table~\ref{tab:reliability_correlation} evaluate whether the phase-derived control signal is directly associated with sample-wise retrieval outcomes. Under both the intra-subject and inter-subject settings, the pooled sample-level correlations between the control signal and retrieval diagnostics remain close to zero, and the subject-wise correlations are likewise centered near zero without a consistent direction across subjects. These results provide little evidence that the current control signal serves as a strong proxy for sample-wise retrieval difficulty. Instead, the evidence is more consistent with interpreting it as an optimization-control cue that modulates supervision strength during training, rather than as a direct estimator of retrieval difficulty.
\begin{table*}[t]
	\centering
	\caption{Correlation analysis between the phase-derived control signal and retrieval diagnostics under intra-subject and inter-subject settings. We report both pooled sample-level correlations and subject-wise mean $\pm$ standard deviation across subjects.}
	\label{tab:reliability_correlation}
	\setlength{\tabcolsep}{5pt}
	\begin{tabular}{llcccc}
		\toprule
		Setting & Metric & Pooled Pearson & Pooled Spearman & Subject-wise Pearson & Subject-wise Spearman \\
		\midrule
		\multirow{3}{*}{Intra-subject}
		& Control signal vs. Margin       & -0.0188 & -0.0388 & $-0.0057 \pm 0.0864$ & $-0.0318 \pm 0.0685$ \\
		& Control signal vs. Negative Rank    &  0.0119 & -0.0323 & $ 0.0449 \pm 0.0673$ & $-0.0268 \pm 0.0631$ \\
		& Control signal vs. Top-1 Correct& -0.0349 & -0.0375 & $-0.0259 \pm 0.0703$ & $-0.0325 \pm 0.0620$ \\
		\midrule
		\multirow{3}{*}{Inter-subject}
		& Control signal vs. Margin       & -0.0030 & -0.0024 & $-0.0032 \pm 0.0676$ & $-0.0075 \pm 0.0671$ \\
		& Control signal vs. Negative Rank    &  0.0049 &  0.0002 & $ 0.0044 \pm 0.0595$ & $-0.0105 \pm 0.0646$ \\
		& Control signal vs. Top-1 Correct& -0.0095 & -0.0058 & $-0.0344 \pm 0.0657$ & $-0.0387 \pm 0.0658$ \\
		\bottomrule
	\end{tabular}
\end{table*}
\section{Ablation Studies}
\label{sec:ablation}
To further understand where the performance gains come from, we conduct a series of ablation studies. We first examine the contribution of the main components, then evaluate generalization across encoder variants, and finally analyze the effects of frequency statistics, temperature control strategies, and the low-dimensional bottleneck size.
\subsection{Component Ablation}
\label{sec:ablation_components}
Table~\ref{tab:ablation_progressive} shows an overall improvement trend as components are progressively introduced, although intermediate variants exhibit metric-specific fluctuations. These results are consistent with the view that the projectors and adapters improve cross-modal conditioning, BWS stabilizes neural embedding geometry, and the addition of frequency-domain modulation together with adaptive temperature further improves retrieval performance under heterogeneous neural conditions.

\begin{table}[t]
	\centering
	\caption{Ablation study of the proposed framework on THINGS-EEG.}
	\label{tab:ablation_progressive}
	\footnotesize
	\setlength{\tabcolsep}{3.5pt}
	\renewcommand{\arraystretch}{1.25}
	\begin{tabular*}{\linewidth}{@{\extracolsep{\fill}}ccccccccc}
		\toprule
		\multicolumn{2}{c}{\textbf{Projectors}} & 
		\multicolumn{2}{c}{\textbf{Adapters}} & 
		\multirow{2}{*}{\textbf{BWS}} & 
		\multirow{2}{*}{\makecell[c]{\textbf{Amp}\\\textbf{Phase}}} & 
		\multirow{2}{*}{\makecell[c]{\textbf{Adaptive}\\\textbf{Temp.}}} & 
		\multirow{2}{*}{\textbf{Top-1}} & 
		\multirow{2}{*}{\textbf{Top-5}} \\
		\cmidrule(lr){1-2} \cmidrule(lr){3-4}
		\textbf{Img} & \textbf{EEG} & \textbf{Img} & \textbf{EEG} & & & & & \\
		\midrule
		$\times$   & $\times$   & $\times$   & $\times$   & $\times$   & $\times$   & $\times$   & 50.15 & 80.00 \\
		\checkmark & $\times$   & $\times$   & $\times$   & $\times$   & $\times$   & $\times$   & 54.70 & 81.60 \\
		$\times$   & \checkmark & $\times$   & $\times$   & $\times$   & $\times$   & $\times$   & 53.35 & 78.55 \\
		\checkmark & \checkmark & $\times$   & $\times$   & $\times$   & $\times$   & $\times$   & 55.10 & 81.65 \\
		\checkmark & \checkmark & \checkmark & \checkmark & $\times$   & $\times$   & $\times$   & 58.25 & 84.75 \\
		\checkmark & \checkmark & \checkmark & \checkmark & \checkmark & $\times$   & $\times$   & 61.45 & \textbf{86.95} \\
		\checkmark & \checkmark & \checkmark & \checkmark & \checkmark & \checkmark & $\times$   & 61.35 & 86.20 \\
		\checkmark & \checkmark & \checkmark & \checkmark & \checkmark & \checkmark & \checkmark & \textbf{63.40} & {86.85} \\
		\bottomrule
	\end{tabular*}
\end{table}

\subsection{Generalization Across Encoder Variants}
\begin{figure}[t]
	\centering
	\subfloat[Top-1 improvement over the UBP.]{
		\includegraphics[width=0.48\textwidth,trim=2mm 0mm 0mm 0mm,clip]{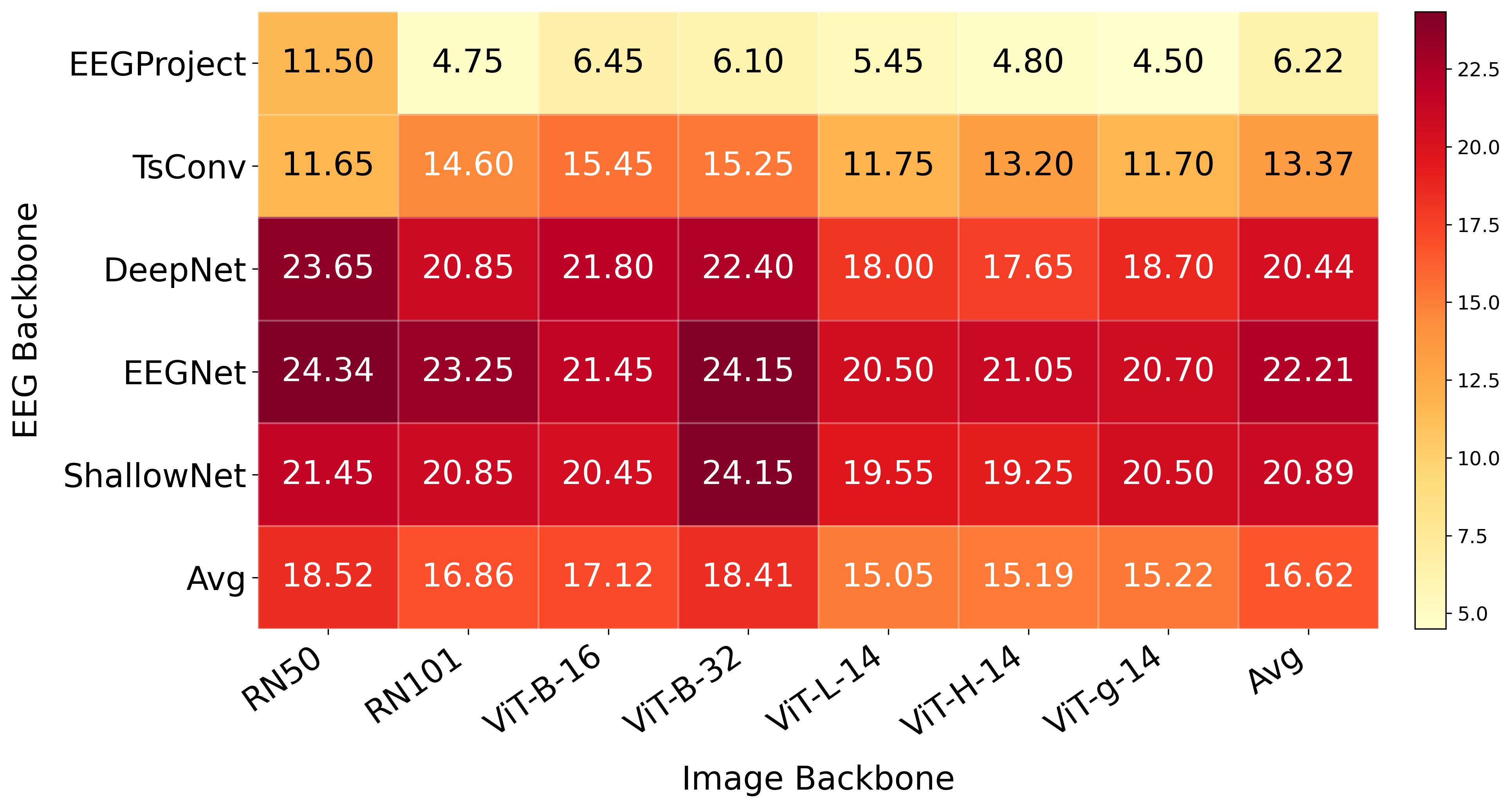}
	}
	\hfill
	\subfloat[Top-5 improvement over the UBP.]{
		\includegraphics[width=0.48\textwidth,trim=2mm 0mm 0mm 0mm,clip]{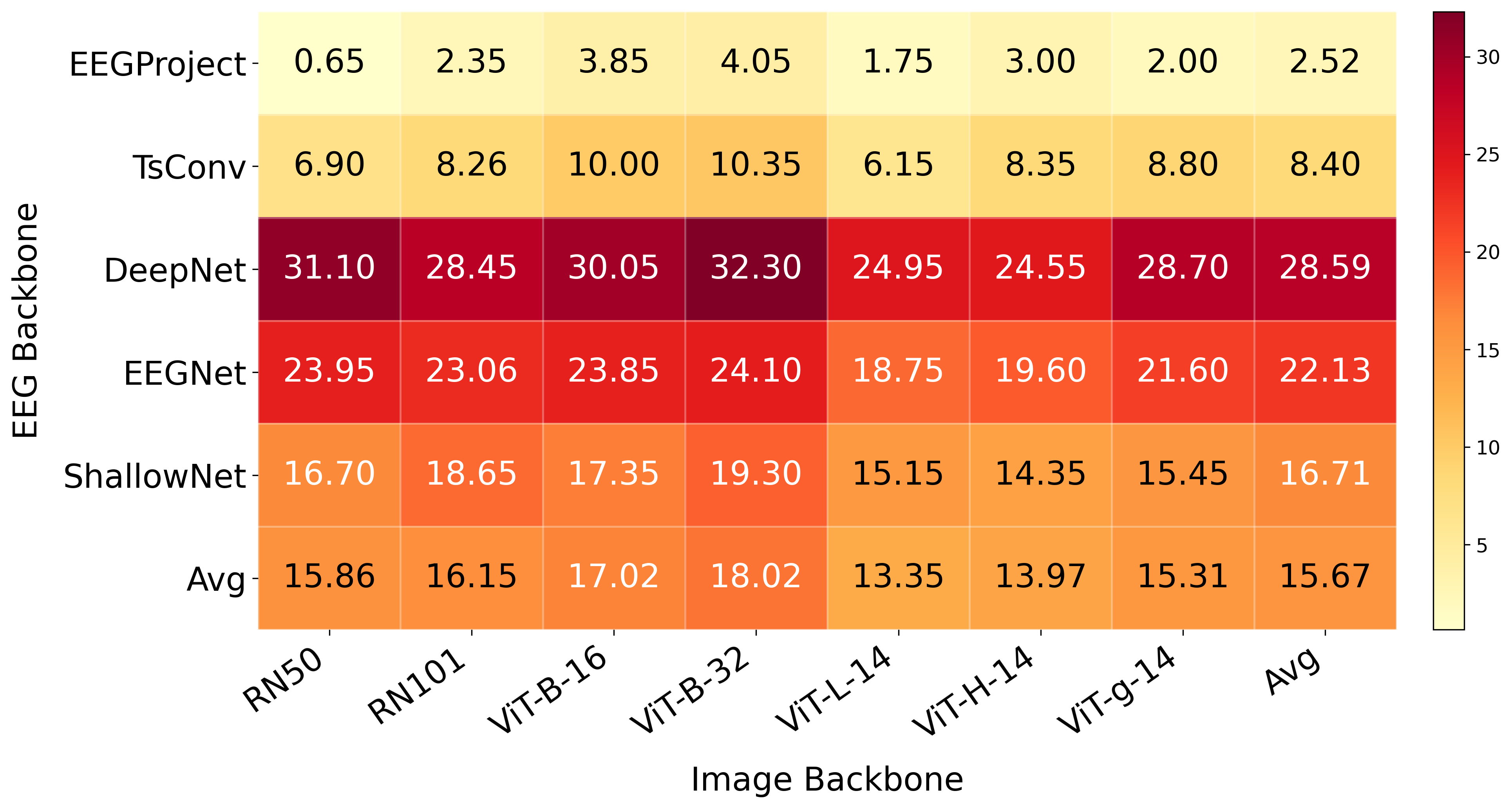}
	}
	\caption{Performance improvement of the proposed method over the UBP baseline across different combinations of EEG backbones and image backbones on THINGS-EEG. Each cell reports the absolute gain (\%) achieved by our method. The last row and last column denote the average improvements across different backbones.}
	\label{fig:ubp_improve_heatmap}
\end{figure}

As shown in Fig.~\ref{fig:ubp_improve_heatmap}, the proposed method improves performance over the UBP baseline across different combinations of EEG and image backbones, indicating that the proposed framework generalizes across both neural and visual encoders. The average gains are more pronounced with RN50, RN101, and ViT-B-32 image backbones, suggesting that the proposed framework is particularly effective under these visual backbones.

\subsection{Ablation of Frequency Statistics and Phase-Guided Adaptive Temperature}

Table~\ref{tab:ablation_freq_temp} reports the ablation results for frequency-based modulation and phase-guided adaptive temperature. Compared with the baseline, the \textit{phase-only} variant yields a larger improvement than the \textit{amplitude-only} variant, suggesting that phase-related statistics may provide more informative cues for alignment in the current retrieval setting. Combining amplitude and phase statistics brings a further gain, indicating that the two types of frequency information are complementary. Adding phase-circular variance improves Top-1 accuracy, although its effect on Top-5 is less consistent. These results indicate that phase-related statistics are more
beneficial to Top-1 retrieval than amplitude-only statistics in the
current setting, while their effect on broader ranking quality is less
consistent. The full model achieves the best Top-1 accuracy,
suggesting that the overall gain is associated with the combination
of frequency-based modulation and adaptive supervision driven by
phase-derived temperature control. In contrast, the best Top-5
performance is achieved by simpler frequency variants, suggesting
that phase-guided adaptive supervision mainly sharpens the
highest-ranked match rather than uniformly improving deeper
ranking positions.
\begin{table}[t]
	\centering
	\caption{Ablation study of frequency-domain statistics and phase-guided adaptive temperature on THINGS-EEG in the intra-subject setting. Amp denotes amplitude statistics, Phase denotes phase statistics, CircVar denotes phase-circular variance, and AdaTemp denotes the phase-guided adaptive temperature mechanism.}
	\label{tab:ablation_freq_temp}
	\begin{tabular}{lccccc}
		\toprule
		Amp & Phase & CircVar & AdaTemp & Top-1 (\%) & Top-5 (\%) \\
		\midrule
		&  &  &  & 61.45 & 86.95  \\
		\checkmark &  &  &  & 61.20 & 86.60  \\
		& \checkmark &  &  & 61.60 & 87.10  \\
		\checkmark & \checkmark &  &  & 61.65 & 87.10 \\
		&  &\checkmark  &\checkmark  & 62.90 & 86.30 \\
		\checkmark & \checkmark & \checkmark &  & 61.35 & 86.20  \\
		\checkmark & \checkmark & \checkmark & \checkmark & 63.40 & 86.85  \\
		\bottomrule
	\end{tabular}
\end{table}
\subsection{Low-Dimensional Ablation}
\begin{figure}[t]
	\centering
	\includegraphics[width=0.9\linewidth]{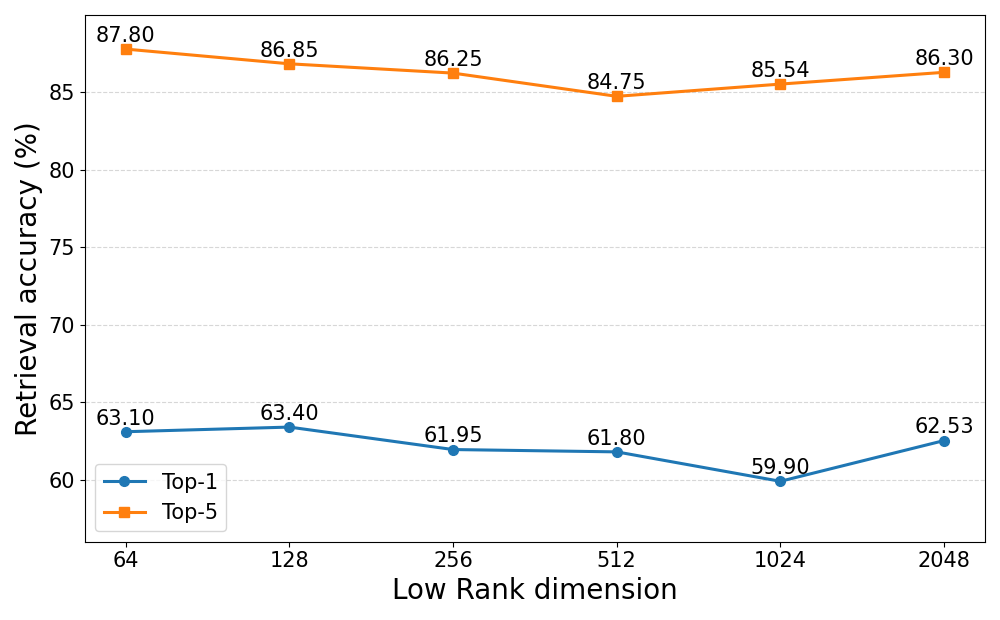}
	\caption{Ablation of the low-dimensional bottleneck size. The model is evaluated with bottleneck sizes ranging from 64 to 2048. The results show that a moderate bottleneck dimension provides a better balance between discriminative capacity and alignment stability, with $r=128$ achieving the best Top-1 performance.}
	\label{fig:lowrank_dim_ablation}
\end{figure}

To further analyze the effect of the bottleneck size in the low-dimensional constraint, we conduct an ablation study on the dimensionality of the shared low-dimensional subspace and evaluate the model with $r \in \{64,128,256,512,1024,2048\}$. As shown in Fig.~\ref{fig:lowrank_dim_ablation}, the effect of the bottleneck dimension is non-monotonic. Very small bottlenecks may limit discriminative capacity by over-compressing cross-modal structure, whereas overly large bottlenecks weaken the structural regularization imposed by the low-dimensional constraint and allow redundant or noisy components to persist. As a result, a moderate bottleneck dimension provides the best trade-off between representational capacity and alignment stability, with $r=128$ achieving the strongest Top-1 performance.

\subsection{Analysis of Temperature Control Strategies}

To evaluate the proposed phase-guided adaptive temperature mechanism, we compare it with three representative alternatives: a fixed temperature, a global learnable temperature, and simple sample weighting. As reported in Table~\ref{tab:temp_mechanism}, the fixed temperature provides a strong baseline, whereas the global learnable temperature performs slightly worse. On this benchmark, a single shared temperature is less effective than sample-aware temperature control. Simple sample weighting yields a modest Top-1 improvement over the fixed-temperature baseline, suggesting that the phase-derived cue may provide optimization-relevant information. In contrast, the proposed phase-guided adaptive temperature achieves the best Top-1 accuracy, while the fixed-temperature baseline achieves the highest Top-5 accuracy. These results suggest that adaptive temperature control mainly sharpens the highest-ranked match in the current setting, rather than uniformly improving all ranking depths.
\begin{table}[t]
	\centering
	\caption{Comparison of different temperature control strategies. Sample-wise indicates whether the temperature or weighting is assigned at the individual-sample level. Phase-guided indicates whether the temperature control is derived from phase stability.}
	\label{tab:temp_mechanism}
	\setlength{\tabcolsep}{6pt}
	\begin{tabular}{lcc}
		\toprule
		Method  & Top-1 (\%) & Top-5 (\%) \\
		\midrule
		Fixed temperature & 63.00 & 87.30 \\
		Global learnable temperature  & 62.40 & 87.15 \\
		Simple sample weighting & 63.15 & 86.55 \\
		Phase-guided adaptive temperature &  63.40 & 86.85 \\
		\bottomrule
	\end{tabular}
\end{table}

\section{Conclusion}
\label{sec:conclusion}
This work presents a unified framework for neural-to-image retrieval that improves cross-modal matching by stabilizing the neural alignment pathway into a frozen visual semantic space. Across THINGS-EEG and THINGS-MEG, the proposed design improves retrieval performance under both intra-subject and inter-subject protocols, highlighting the value of jointly integrating representation stabilization, frequency-guided calibration, and adaptive supervision for robust neural-to-image retrieval.

\textbf{Limitations}: First, the current frequency branch relies on relatively compact statistical descriptors and therefore does not fully capture richer time-frequency structure or finer-grained neural dynamics. Second, although the proposed framework is motivated and analyzed from the perspectives of representational stability and alignment robustness, the reason why the frequency-domain enhancement is especially beneficial for more challenging samples still requires deeper neuroscientific interpretation and stronger statistical validation. Third, the phase-derived control signal should be interpreted primarily as an optimization-control cue rather than as a direct estimator of sample-wise retrieval difficulty, and its relationship to underlying neural reliability remains to be clarified further.

\bibliographystyle{IEEEtran}
\bibliography{references}

\end{document}